\documentclass[journal]{IEEEtran}

\usepackage{amsmath}
\usepackage{graphicx}
\usepackage{subfig} 
\usepackage{tabu}  

\usepackage[colorlinks,linkcolor=blue,anchorcolor=blue,citecolor=green]{hyperref}
\usepackage{multirow}
\usepackage{dsfont}
\usepackage{stfloats}
\usepackage{amsmath}
\usepackage{cite}
\usepackage{booktabs}
\usepackage{algorithm}
\usepackage{algorithmic}
\usepackage{amssymb}
\ifCLASSINFOpdf

\else
 
\fi

\begin{document}

\title{DSDANet: Deep Siamese Domain Adaptation Convolutional Neural Network for Cross-domain Change Detection}

\author{
        Hongruixuan~Chen,~\IEEEmembership{Student~Member,~IEEE,}  
        Chen~Wu,~\IEEEmembership{Member,~IEEE,}      
        Bo~Du,~\IEEEmembership{Senior~Member,~IEEE,}
        and~Liangpei~Zhang,~\IEEEmembership{Fellow,~IEEE}% <-this % stops a space

\thanks{Manuscript submitted June 15, 2020. This work was supported in part by the National Natural Science Foundation of China under Grant 61971317, 41801285, 61822113 and 41871243.}
\thanks{H. Chen is with the State Key Laboratory of Information Engineering in Surveying, Mapping and Remote Sensing, Wuhan University, Wuhan, P.R. China (e-mail:Qschrx@whu.edu.cn).}
\thanks{C. Wu is with the State Key Laboratory of Information Engineering in Surveying, Mapping and Remote Sensing, and School of Computer Science, Wuhan University, Wuhan, P.R. China (e-mail: chen.wu@whu.edu.cn, corresponding author).}% <-this % stops a space
\thanks{B. Du is with the School of Computer Science, and Collaborative Innovation Center of Geospatial Technology, Wuhan University, Wuhan, P.R. China (email: gunspace@163.com).}% <-this % stops a space
\thanks{L. Zhang is with the Remote Sensing Group, State Key Laboratory of Information Engineering in Surveying, Mapping, and Remote Sensing, Wuhan University, Wuhan, P.R. China (e-mail: zlp62@whu.edu.cn).}
}

% The paper headers
\markboth{SUBMITTED TO IEEE TGRS ON June 15, 2020}%
{Shell \MakeLowercase{\textit{et al.}}: Bare Demo of IEEEtran.cls for IEEE Journals}

% make the title area
\maketitle

% As a general rule, do not put math, special symbols or citations
% in the abstract or keywords.
\begin{abstract}
  Change detection (CD) is one of the most vital applications in remote sensing. Recently, deep learning has achieved promising performance in the CD task. However, the deep models are task-specific and CD data set bias often exists, hence it is inevitable that deep CD models would suffer degraded performance after transferring it from original CD data set to new ones, making manually label numerous samples in the new data set unavoidable, which costs a large amount of time and human labor. How to learn a transferable CD model in the data set with enough labeled data (original domain) but can well detect changes in another data set without labeled data (target domain)? This is defined as the cross-domain change detection problem. In this paper, we propose a novel deep siamese domain adaptation convolutional neural network (DSDANet) architecture for cross-domain CD. In DSDANet, a siamese convolutional neural network first extracts spatial-spectral features from multi-temporal images. Then, through multi-kernel maximum mean discrepancy (MK-MMD), the learned feature representation is embedded into a reproducing kernel Hilbert space (RKHS), in which the distribution of two domains can be explicitly matched. By optimizing the network parameters and kernel coefficients with the source labeled data and target unlabeled data, DSDANet can learn transferrable feature representation that can bridge the discrepancy between two domains. To the best of our knowledge, it is the first time that such a domain adaptation-based deep network is proposed for CD. The theoretical analysis and experimental results demonstrate the effectiveness and potential of the proposed method. 
\end{abstract}

% Note that keywords are not normally used for peerreview papers.
\begin{IEEEkeywords}
  Change detection, deep learning, transfer learning, domain adaptation, convolutional neural network (CNN), multi-kernel maximum mean discrepancy (MK-MMD), multispectral images.
\end{IEEEkeywords}

\IEEEpeerreviewmaketitle

\section{Introduction}\label{sec:1}

\IEEEPARstart{D}{ue} to the rapid development of Earth observation technology, a large number of multi-temporal remote sensing images covering the same area are available now. How to utilize these massive data to obtain accurate and real-time land surface changes is significant for a better understanding of human activities, ecosystem, and their interactions \cite{Coppin2004}. Therefore, change detection (CD) has become one of the hot topics in the remote sensing field, which is defined as the process of identifying differences in the state of an object or phenomena by observing it at different times \cite{Singh1989}. Now, CD has been widely applied to many real-world applications, such as ecosystem monitoring, resource management, urban expansion research, and damage assessment \cite{Lu2004, Brunner2010, Deng2008, Xian2009, Zelinski2014, Luo2018, Qin2016, Wu2018}. 

\par Consequently, numerous kinds of conventional CD models have been proposed in the literature. These approaches can be concluded as the following main categories: 1) Image algebra. Such methods do algebraic operation band to band directly, including image difference method, image ratio method, image regression method, and change vector analysis (CVA) \cite{Sharma2007}, which has been the basis of some more advanced CD models \cite{Thonfeld2016, Saha2019, Du2020}. 2) Image transformation models transform images into new feature space, where change information can be highlighted. Some typical methods are multivariate alteration detection (MAD) \cite{Nielsen1997, Nielsen2007, Volpi2015}, principal component analysis (PCA) \cite{Deng2008, Celik2009}, and slow feature analysis (SFA) \cite{Wu2014, Wu2017a}. 3) Classification methods mainly indicate post-classification comparison \cite{Wu2017b} and compound classification approach \cite{Bovolo2008}. These methods can provide detailed “from-to” information about land-cover and land-use changes \cite{Wu2017a}. 4) Other advanced models——including some probability graph models, such as Markov random filed \cite{Benedek2009} and conditional random filed \cite{Hoberg2015, Lv2018a}, wavelet \cite{Celik2011}, texton forest \cite{Lv2018a}, object-based models \cite{Desclee2006, Hussain2013, Jabari2019, Gil-Yepes2016}, and so on. However, these CD models only explore “shallow” features and many of them are hand-crafted, which is insufficient for representing the key information of original data and limits their performance.

\par Recently, deep learning (DL) has been shown to be very promising in the field of computer vision and remote sensing images interpretation \cite{Zhang2016b, Ma2019}. In comparison with the aforementioned conventional models, DL models are capable of extracting representative deep features from multispectral images. Hence, a lot of CD methods based on DL models are developed. By constructing a deep neural network (DNN), in \cite{Gong2016}, an efficient CD method is proposed for multi-temporal SAR. In \cite{Gong2017a}, Gong et. al design a hybrid deep CNN to learn representative features for ternary CD.  Lyu et al. \cite{Lyu2016} propose an efficient CD method based on recurrent neural network (RNN). In this work, an improved long-short term memory (LSMT) is adopted to learn transferable change rules for multispectral images. By introducing siamese network into CD, Zhan et al. \cite{Zhan2017} present a deep siamese convolutional network for CD in aerial images, which extracts deep spatial-spectral features by two weight-shared branches. Aiming to integrate the merits of both CNN and RNN to learn spatial-spectral-temporal features, Mou et al. \cite{Mou2019} first adopt recurrent convolutional neural network architecture for CD in multispectral images. Going one step further, Chen et al. \cite{Chen2019a} specifically design a deep siamese convolutional multiple-layer recurrent neural network (SiamCRNN) for multisource CD. In SiamCRNN, a deep siamese CNN is designed for extracting deep spatial-spectral features and a multiple-layer RNN is adopted for mapping spatial-spectral features into a new latent feature space and mining change information. Combining a dual-streams deep neural network (DNN) and SFA, Du et al \cite{Du2019a} present a unified model called deep slow feature analysis (DSFANet) for multispectral CD. In \cite{CayeDaudt2018}, fully convolutional network (FCN) is first introduced into multispectral CD and three FCN architectures are designed by Daudt et. al. After that, Chen et. al \cite{Chen2019} adopt a multi-scale feature convolution unit to learn multi-scale spatial-spectral features. Based on the unit, they design a deep siamese multi-scale FCN to achieve a better CD result on an open CD data set. To better extract joint spatial-spectral features, through extending 2-D convolution to 3-D convolution, Song et. al \cite{Song2018} propose a recurrent 3D FCN for binary and multi-class CD. 

\par Nonetheless, the training procedure of these DL-based CD methods inevitably requires a lot of labeled data and there is no denying that the manual selection of labeled data is labor-consuming, especially for labelling changes in multi-temporal remote sensing images. More importantly, deep networks are often task-specific, in other words, they have a relatively weak generalization \cite{Yosinski2014}. And due to several factors, such as noise and distortions, sensor characteristics, imaging conditions, the data distributions of different CD data sets are often quite dissimilar. Thus, if we train a deep network on one CD data set with abundant labeled samples, it would suffer degraded performance after we transfer it to a new CD data set, which makes it unavoidable to manually label numerous samples in the new data set. Consequently, in consideration of fully utilizing the trained model to reduce the training cost and obtaining promising CD results in the new data sets, it is an incentive to develop an efficient CD model that is trained on a data set (source domain) with enough labeled data but can be easily transferred to a new data set (target domain) with no (or very limited) labeled data. This we called as the cross-domain change detection problem in this paper. To solve the cross-domain CD problem, a few works try to utilize transfer learning techniques, such as \cite{Yang2019a, Song2020, Song2020a}, by just adopting the standard fine-tuning strategy. However, such a standard fine-tuning still requires sufficient labeled data in the target domain. Besides, fine-tuning assumes that the data distributions of two data sets are similar \cite{Tzeng2015}. If the domain distribution discrepancy is great, the performance of fine-tuning is seriously degraded. 

\par Considering the above issues comprehensively, in this paper, a novel deep siamese domain adaptation convolutional neural network (DSDANet) is proposed for cross-domain CD. By incorporating a domain discrepancy metric MK-MMD into the network architecture and optimizing the network parameters and kernel coefficient together, DSDANet can learn transferrable features, where the distribution of two domains would be similar. Thus, only requiring very sparse labeled data for target domain, DSDANet can achieve promising performance. The contributions of this paper can be concluded as follows:

\begin{enumerate} 
  \item To efficiently solve the cross-domain CD problem, we propose a novel deep network architecture called DSDANet. By jointly learning representative deep features and minimizing domain discrepancy, DSDANet can learn transferable features. To the best of authors’ knowledge, it is the first time that such a deep network based on domain adaptation is designed for CD in multispectral images, which explores a new perspective for cross-domain CD.
  \item In DSDANet, for enhancing the transferability of difference feature to cope with some situations of severe domain distribution discrepancy, a multi-kernel MMD with multiple-layer domain adaptation schema is adopted. 
  \item We conduct three detailed and persuasive cross-domain CD experiments on four CD data sets. The experimental results demonstrate that our proposed DSDANet is capable of efficiently minimizing the distance between two domains and learning transferable features, leading to competitive CD performance, particularly in the situation of large domain distribution discrepancy. 
\end{enumerate}

\par The rest of this paper \footnote{This paper is an improved version of the original paper: https://arxiv.org/abs/2004.05745} is organized as follows. In section \ref{sec:2}, the background of CNN, domain adaptation, and cross-domain CD are introduced. Section \ref{sec:3} elaborates MK-MMD and the proposed DSDANet. To evaluate the proposed model, the experiments of the cross-domain CD are carried in sections \ref{sec:4}. Finally, Section \ref{sec:5} draws the conclusion of our work in this paper.
% You must have at least 2 lines in the paragraph with the drop letter
% (should never be an issue)

\section{Background}\label{sec:2}
\subsection{CNN}
\par In deep learning, CNN is a class of deep models designed for processing data that come in the form of multiple arrays, such as audio spectrograms, images, and video. Now, owing to the advantages of sparse interactions, parameter sharing, and equivariant representations, CNN has been widely employed in the computer vision tasks and has achieved promising performance \cite{Lecun2015}.

\par Generally, CNN consists of three basic kinds of layers, namely, convolutional layer, pooling layer, and fully connected (FC) layer. Among them, convolutional layer convolves input data with some trainable convolution kernels to extract a feature map, i.e. $F=C\left(I\right)$
\begin{equation}  
  F^{(i)}=C^{(i)}\left(I\right)=a\left(W^{\left(i\right)}\ast I+b^{\left(i\right)}\right)
\end{equation}
where $F^{(i)}$ is the i-th channel of the extracted feature map, $W^{(i)}$ is the i-th convolution kernel, $b^{(i)}$ is its corresponding bias, $a\left(\cdot\right)$ is an activation function, which can introduce non-linearity into the extracted features. Pooling layer replaces the feature map at a certain location with a summary statistic of the nearby outputs, which helps the feature map become approximately invariant to small translations of the input, enlarge the receptive field of networks, and mitigate overfitting. FC layer integrates features and mapping them into the target label space
\begin{equation}  
  y=a\left(Wf+b\right)
\end{equation}
where $y$ is the predicted label, $W$ and $b$ are weight and bias,  $a$ is an activation function. 

\par A typical CNN architecture often first extracts high-level abstract features with a series of convolutional layers and pooling layers, then predicts the final desired result with one or several FC layers. By designing a loss function $\mathcal{L}$ measuring the discrepancy between the predicted result $Y$ and true result $\hat{Y}$, the optimization of CNN can be formulated as $\min\limits_{\Theta}\mathcal{L}\left(Y,\hat{Y}\right)$. Here, $\Theta$ is the trainable parameters of CNN. To optimize this objective function, the most commonly used algorithm is the back-propagation (BP) algorithm \cite{Y1998}. 

\subsection{Domain Adaptation}
\par Domain adaptation is a special case of transfer learning. In domain adaptation, a domain $\mathcal{D}$ is composed of a feature space $\mathcal{X}$ and possibility distribution $P\left(X\right)$, here $X=\left\{x_{1},x_{2},\cdots,x_{n}\right\}\in\mathcal{X}$ is the set of data. A task $\mathcal{T}$ has two parts: a label space $\mathcal{Y}$ and a predictive function $f\left(\cdot\right)$. Specifically, the domain with sufficient labeled data is called source domain, denoted as $\mathcal{D}_{s}=\left\{X_{s},Y_{s}\right\}=\left\{\left(x_{s_{i}},y_{s_{i}}\right)\right\}_{n=1}^{n_{s}}$, where $x_{s_{i}}\in\mathcal{X}_{s}$ is the i-th data instance and $y_{s_{i}}\in\mathcal{Y}_{s}$ is the corresponding label. The domain without labeled data is called target domain, denoted as $\mathcal{D}_{s}=\left\{X_{t},Y_{t}\right\}=\left\{\left(x_{t_{i}},y_{t_{i}}\right)\right\}_{n=1}^{n_{t}}$, here, $y_{t_{i}}\in\mathcal{Y}_{t}$ is generally not given. Assuming that the source domain $\mathcal{D}_{s}$ and target domain $\mathcal{D}_{t}$ have the same feature space and label space, i.e. $\mathcal{X}_{s}=\mathcal{X}_{t}$ and $\mathcal{Y}_{s}=\mathcal{Y}_{t}$, but marginal distributions of the two domains are different, $P\left(X_{s}\right)\neq P\left(X_{t}\right)$, domain adaptation aims to improve the performance of $f_{\mathcal{D}_{t}}\left(\cdot\right)$ in $\mathcal{D}_{t}$ by using the knowledge of $\mathcal{D}_{s}$. A commonly adopted way in domain adaptation is to find a mapping $\Phi\left(\cdot\right)$ that maps data of two domains into a new common feature space, where the data distribution of two domains could be explicitly aligned.

\begin{figure}[!t]

  \centering
  \subfloat[]{
    \includegraphics[width=3.3in]{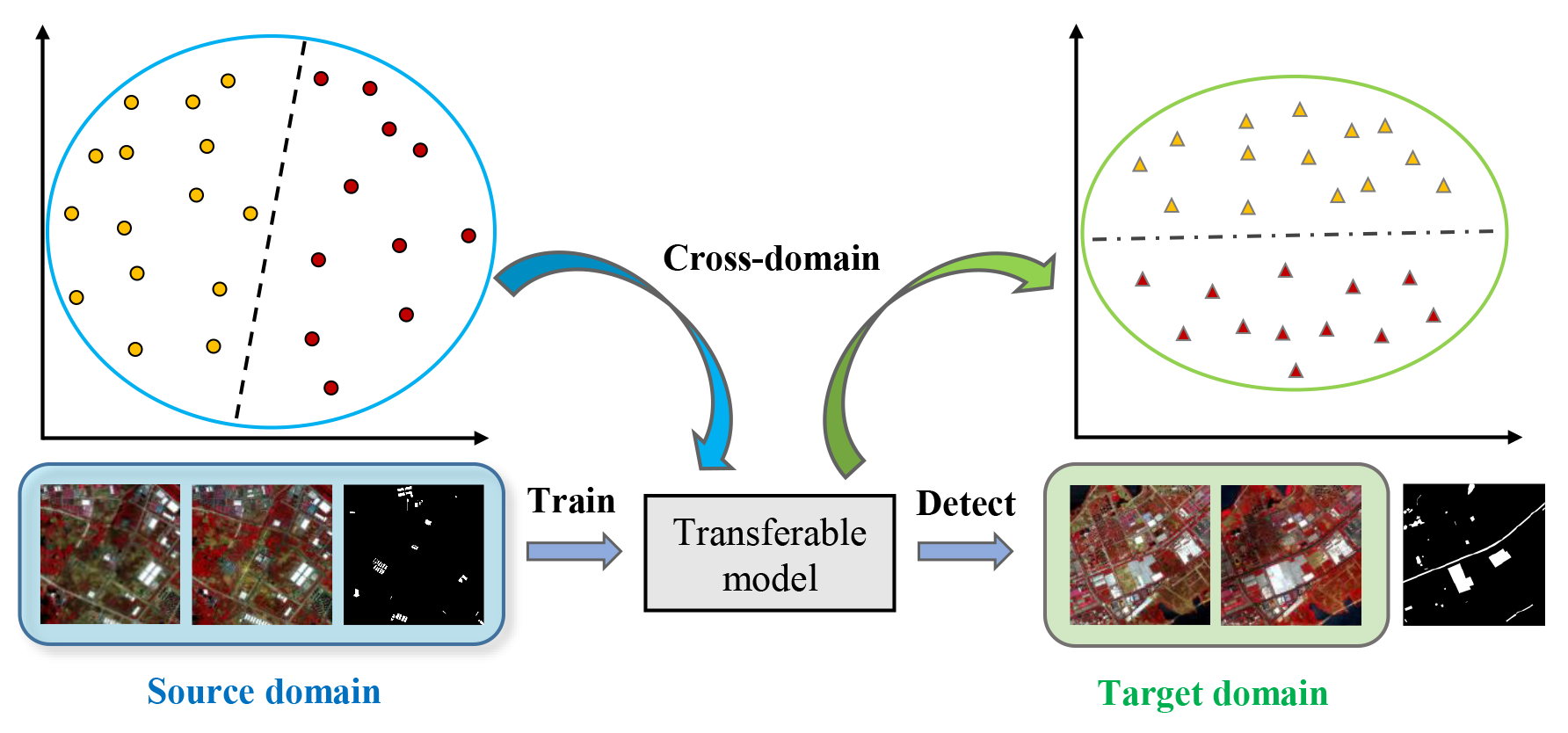}
  \label{fig:CSCD}}

  \subfloat[]{
    \includegraphics[width=3.3in]{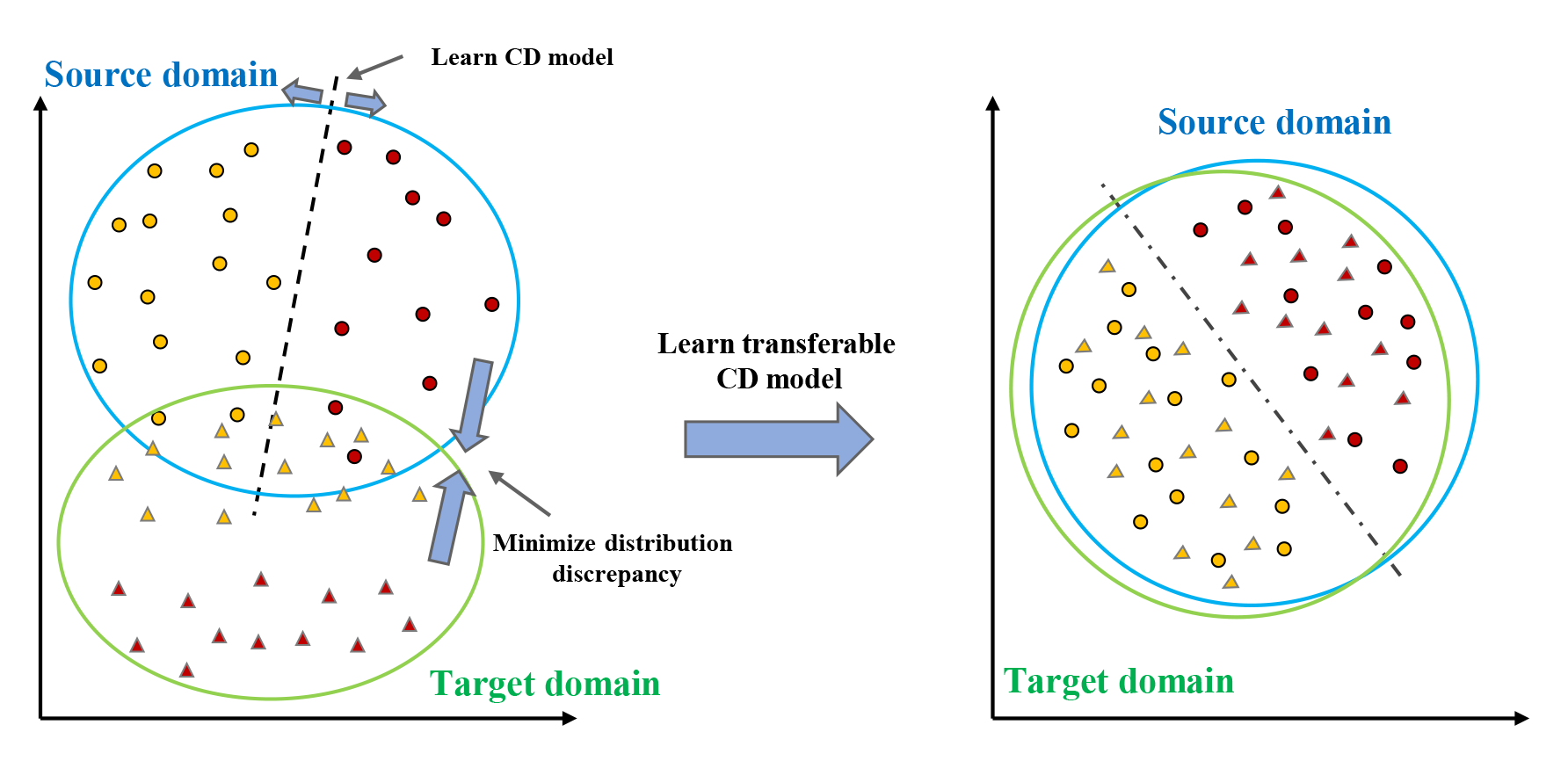}
  \label{fig:moti}}
  \caption{Illustration of (a) cross-domain CD and (b) our motivation.}

  \label{fig:CSCD}
\end{figure}
\subsection{Problem Statement}
\par Suppose a CD data set (source domain) $\mathcal{D}_{s}=\left\{X_{s},Y_{s}\right\}=\left\{X_{s}^{T_{1}},X_{s}^{T_{2}},Y_{s}\right\}$ with sufficient labeled data and another data set (target domain) $\mathcal{D}_{s}=\left\{X_{t}\right\}=\left\{X_{t}^{T_{1}},X_{t}^{T_{2}}\right\}$ with no labeled data, caused by plenty of factors, the data distributions of the two data sets are dissimilar $P\left(X_{s}\right) \neq P\left(X_{t}\right)$. How to learn a robust and transferable model $f\left(\cdot\right)$ in the $D_{s}$ with enough labeled data but can well detect changes in the $D_{t}$ providing no labeled data? In this paper, we call it as cross-domain CD, as shown in Fig. \ref{fig:CSCD}-(a). Furthermore, according to the difficulty of tasks, cross-domain CD can be cataloged as the following three cases. 
\begin{enumerate} 
  \item First is cross-domain with single source data sets, which is also the most common scenery, the source domain and target domain are captured by the same sensor, but they covered different areas with different ground objects distribution and imaging condition. 
  \item The more difficult case is cross-domain with multi-source but homogeneous data sets. For instance, the target data set is captured by GF-2 but the target data set is acquired by Quickbird. Compared with the case 1, the distribution discrepancy between two domains in case 2 is greater. 
  \item Last and the most challenging, cross-domain with multi-source and heterogeneous data sets, the data in the source domain and the data in the target domain are acquired by different types of sensors. In this case, the statistical characteristic and the data distributions of two domains are totally different, the data of two domains is located in different feature space, $\mathcal{X}_{s} \neq \mathcal{X}_{t}$. For example, the source domain is multispectral but the target domain is In-SAR or LiDAR data.  
\end{enumerate}

\par In this paper, we mainly focus on the first two cases and multispectral data. What’s more, for the target domain $D_{t}$, we extend the condition of no labeled data to the existence of very limited labels $\left\{y_{t_{i}}\right\}_{i=1}^{n_{t}^{\prime}}$, where $n_{t}^{\prime}\ll n_{s}$ and $n_{t}^{\prime}\ll n_{t}$. 

\par To solve the cross-domain CD problem, as illustrated in Fig. \ref{fig:CSCD}-(b), our motivation is that when the deep model learns representative features in the source labeled data, the distribution discrepancy of two domains can also get minimized, in that way, the model is capable of learning transferable representation, hence can be easily transferred from one CD data set to another.

\section{Methodology}\label{sec:3}
\subsection{MK-MMD}
\par Since our motivation is jointly learning representative features and restricting the domain distribution discrepancy, we first need to find an appropriate metric to represent the domain distribution discrepancy. 

\par In domain adaptation, a widely used metric is the maximum mean discrepancy (MMD) \cite{Borgwardt2006}. MMD is a nonparametric kernel-based metric that measures the distance between two distributions in an RKHS. Given a source domain $D_{s}=\left\{X_{s},Y_{s}\right\}=\left\{\left(x_{s_{i}},y_{s_{i}}\right)\right\}_{i=1}^{n_{s}}$ and a target domain $D_{t}=\left\{X_{s}\right\}=\left\{x_{s_{i}}\right\}_{i=1}^{n_{t}}$, the empirical estimate of the MMD between the two domains’ distributions can be expressed as follows:
\begin{equation}  
  d\left(X_{s}, X_{t}\right)=\left\|\frac{1}{n_{s}} \sum_{i=1}^{n_{s}} \Phi\left(x_{s_{i}}\right)-\frac{1}{n_{t}} \sum_{j=1}^{n_{t}} \Phi\left(x_{t_{j}}\right)\right\|_{\mathcal{H}}
\end{equation}

where $\left\|\cdot\right\|_{\mathcal{H}}$ is the RKHS norm and $\Phi\left(\cdot\right)$ is the kernel-induced feature map. On the contrary, if the distributions of two domains tend to be the same and the RKHS is universal, MMD would approach zero. On the other contrary, if a model can learn a domain-invariant representation that minimizes the MMD between two domains, it can be easily transferred to the target domain with limited labeled data.  

\par Nonetheless, it is difficult to find an optimal RKHS and the representation ability of a single kernel is limited. And it is reasonable to assume that the optimal RKHS can be expressed as the linear combination of single kernels, thus the multi-kernel variant of MMD entitled MK-MMD \cite{Gretton2012} is introduced.

\begin{figure*}[!t]

  \centering
  \includegraphics[scale=0.45]{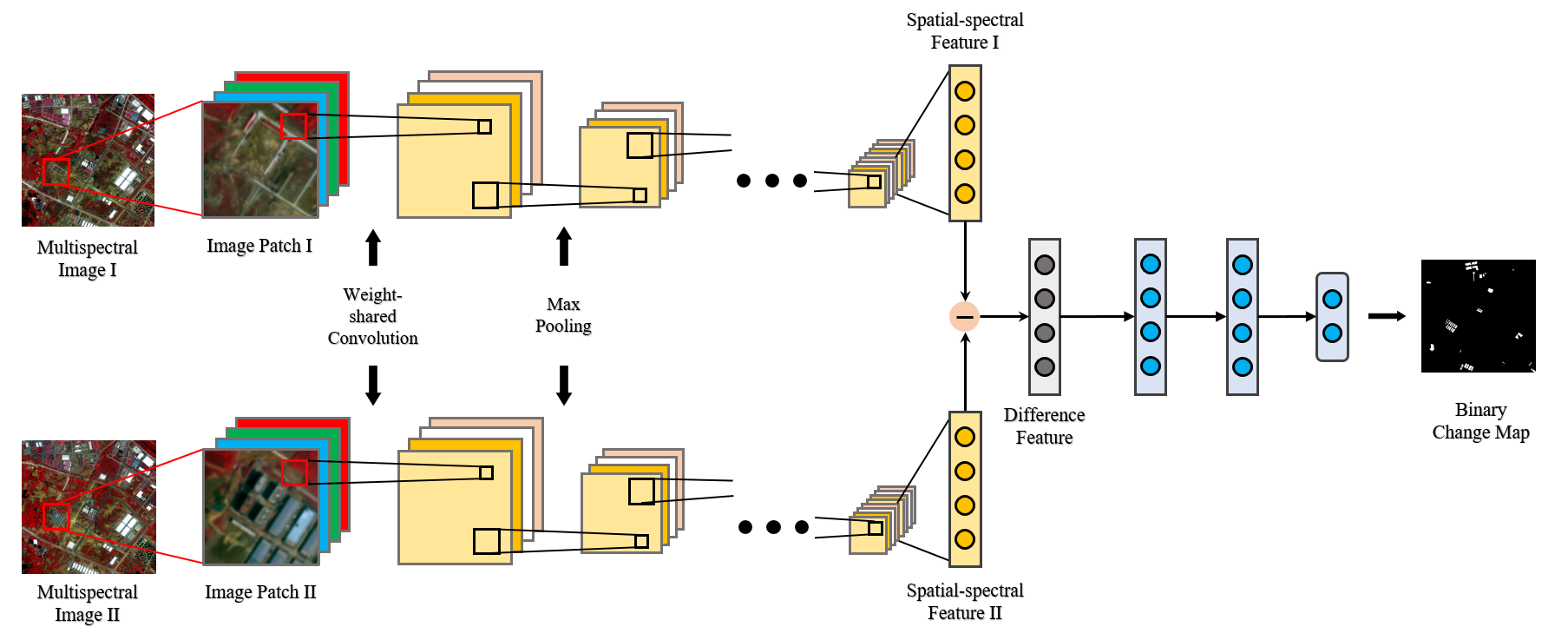}
  \caption{Illustration of DSCNet. DSCNet extracts spatial-spectral features from multispectral image patches with two siamese convolutional streams, through subtraction and taking absolute values, difference feature is calculated, finally several FC layers further transform difference feature to get the final CD result.}
  \label{fig:DSCNet}
\end{figure*}

\par Considering two domains $\mathcal{D}_{s}$ and $\mathcal{D}_{t}$, the formulation of MK-MMD is defined as
\begin{equation}
  d_{k}\left(X_{s}, X_{t}\right)=\left\|E\left(\Phi_{k}\left(X_{s}\right)\right)-E\left(\Phi_{k}\left(X_{t}\right)\right)\right\|_{\mathcal{H}}
\end{equation}
where $\Phi_{k}\left(\cdot\right)$ is the feature map induced by multi-kernel $\mathcal{K}$, which is defined as the linear combination of $n$ positive semi-definite kernels $\left\{k_{u}\right\}_{u=1}^{n}$ 
\begin{equation}
  \begin{split}
    \mathcal{K}:=&\left\{k: k=\sum_{u=1}^{n} \beta_{u} k_{u}, \right. \\ 
    & \ \ \ \left. \sum_{u=1}^{n} \beta_{u}=1, \beta_{u} \geq 0, u \in\{1,2, \ldots, n\}\right\} 
  \end{split}
\end{equation}
where each $k_{u}$ is associated uniquely with an RKHS $\mathcal{H}$, and we assume the kernels are bounded. Owing to leveraging diverse kernels, the representation ability of MK-MMD can get improvement. Cooperating the MK-MMD into a deep model, the DSDANet is designed for cross-domain CD in multispectral images. 

\subsection{Network Architecture}
\subsubsection{DSCNet}
\par The backbone of our proposed method is a deep siamese convolutional neural network (DSCNet). As shown in Fig. \ref{fig:DSCNet}, same with the typical CNN architecture, DSCNet consists of convolutional layers, pooling layers and FC layers. Given an image patch-pair $\left(x_{i}^{T_{1}},x_{i}^{T_{2}}\right)$, $x_{i}^{T_{n}}\in \mathcal{R}^{k_{1}\times k_{2} \times c}$ is an image patch centered i-th pixel, the deep spatial-spectral features are extracted by cascade convolutional layers and pooling layers, this procedure can be expressed as
\begin{equation}
  \left\{\begin{array}{l}
    f_{i}^{T_{1}}=S_{1}\left(x_{i}^{T_{1}}\right)=P_{1}^{L_{1}}\left(C_{1}^{L_{1}} \cdots C_{1}^{1}\left(x_{i}^{T_{1}}\right)\right) \\
    f_{i}^{T_{2}}=S_{2}\left(x_{i}^{T_{2}}\right)=P_{2}^{L_{1}}\left(C_{2}^{L_{1}} \cdots C_{2}^{1}\left(x_{i}^{T_{2}}\right)\right)
    \end{array}\right.
\end{equation}
Here, $S_{1}$ and $S_{2}$ are the two siamese convolutional streams of DSCNet, $L$ is the stream depth, $C$ and $P$ denote the pooling layer and convolutional layer, respectively. After that, the absolute value of the difference between deep spatial-spectral features $f_{i}^{T_{1}}$ and $f_{i}^{T_{2}}$ is calculated $d_{i}=\left|f_{i}^{T_{1}}-f_{i}^{T_{2}}\right|$. Since $S_{1}$ and $S_{2}$ are weight-shared, change information could be highlighted through this operation \cite{Liu2018, Zhang2019c, Chen2019}. After generating difference features, several FC layers further mine the underlying change information in difference features to get the final CD result
\begin{equation}
  \begin{split}
    p_{i}=&FC(d_{i}) \\
    =&a^{L}\left(W^{L} a^{L-1}\left(\cdots a^{1}\left(W^{1} d_{i}+b^{1}\right)\right)+b^{L}\right)
  \end{split}
\end{equation} 
where $p_{i}$ is $i$-th pixel change probability, $L$ is layer numbers of FC, $W^{l}$ and $b^{l}$ are parameters of $l$-th layer, and $a^{l}$ is activation function of $l$-th layer. 

\begin{figure*}[!t]

  \centering
  \includegraphics[scale=0.23]{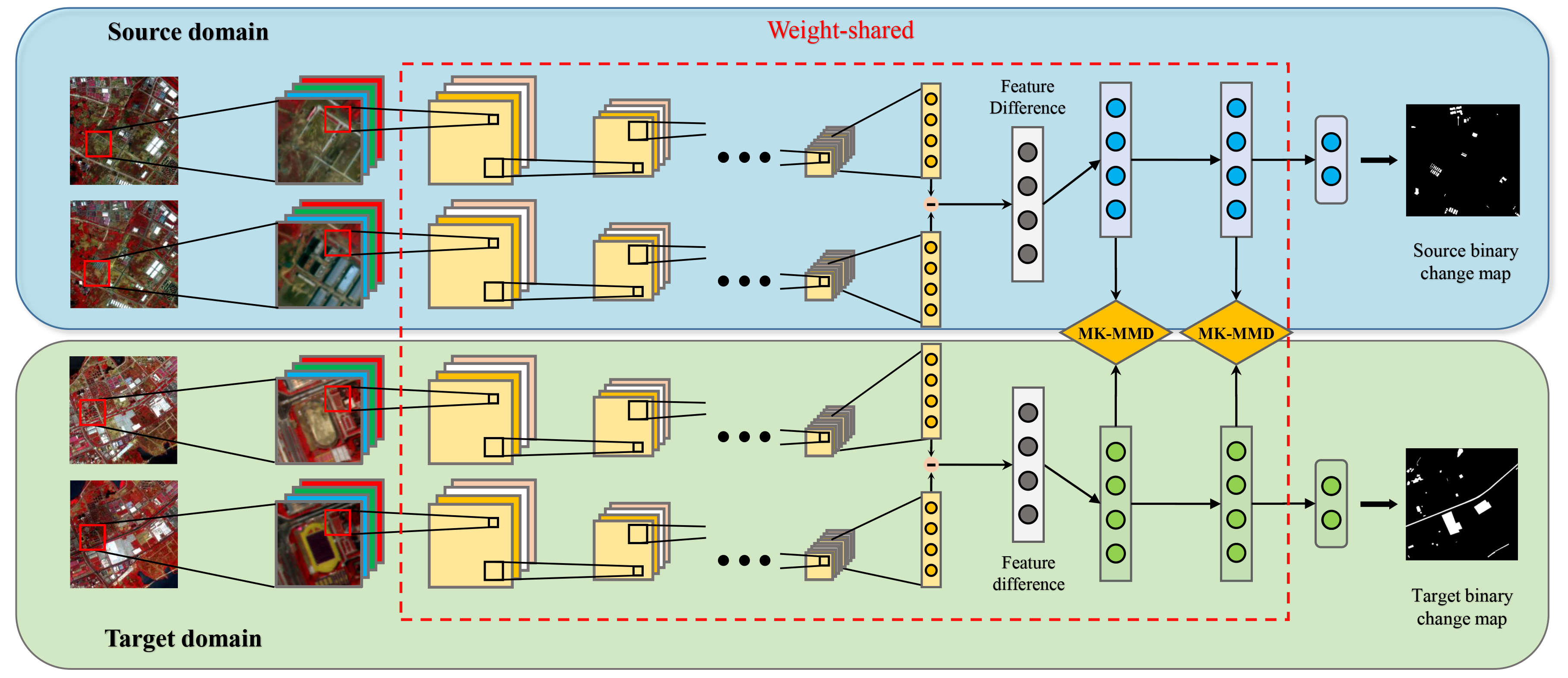}
  \caption{Overview of the CD architecture based on the proposed DSDANet. In the training procedure, DSDANet learns representative difference features on the source labeled data, meanwhile, by minimizing the domain discrepancy measured by MK-MMD, DSDANet is capable of bridging the discrepancy between two domains, so as to learn transferable features.}
  \label{fig:DSDANet}
\end{figure*}

\subsubsection{DSDANet}
\par DSCNet is designed for conventional multispectral CD tasks and not applicable for cross-domain CD. As observed by Yosinski et. al \cite{Yosinski2014}, deep features learned by CNN transition from general to specific by the network going deeper. Especially for the last few FC layers, there exists an insurmountable transferability gap between features learned from different domains. Directly fine-tuning is a seems feasible strategy. Nonetheless, fine-tune still requires sufficient labeled data in the target domain. Besides, the basic assumption of fine-tune is the data distributions of two data sets is similar. Thus, if the distribution difference is great, the effect of fine-tune is seriously damaged (this is also verified in section 4.4). 

\par Our motivation is that if we can learn a feature representation that minimizes the distance between the source and target distributions, then we can train a transferable deep model on the source labeled data and directly apply it to the target domain with very limited labeled data. However, due to $P\left(X_{s}\right) \neq P\left(X_{t}\right)$, we cannot minimize the distance between the two domains directly. Therefore, the MK-MMD is adopted to make the DSCNet learn transferable difference features from two domains.

\par The remaining question is choosing where in the network to embed the MK-MMD. A preliminary idea is combining MK-MMD with the penultimate FC layer because the difference features of this layer is most tailored to fit specific-tasks and most difficult to transfer, thus should be adapted with MK-MMD, which can directly make the classifier adaptive to two domains \cite{Tzeng2015}. But considering a single layer adaptation is insufficient to cope with some situations of severe domain distribution bias, especially for the case 2 of cross-domain CD, hence we adopt a multiple-layer adaptation schema \cite{Long2015} that the MK-MMD is embedded into the two FC layers in front of the classifier. The FC layer embedding with MK-MMD is called as adaptation layer. 

\par Introducing MK-MMD into DSCNet for domain adaptation, the architecture of our proposed DSDANet is shown in Fig. \ref{fig:DSDANet}. Given a source domain $D_{s}=\left\{X_{s}, Y_{s}\right\}=\left\{\left(x_{s_{i}}^{T_{1}}, x_{s_{i}}^{T_{2}}, y_{s_{i}}\right)\right\}_{i=1}^{n_{s}}$ with enough labeled data and a target domain $D_{t}=\left\{X_{t}\right\}=\left\{\left(x_{t_{i}}^{T_{1}}, x_{t_{i}}^{T_{2}}\right)\right\}_{i=1}^{n_{t}}$ without labeled data, since we aim to construct a network that is trained on the source CD data set but also perform well on the target domain, the loss function of DSDANet is designed as
\begin{equation}
  \mathcal{L}=\mathcal{L}_{C}\left(X_{s}, Y_{s}\right)+\lambda \sum_{l=l_{a}}^{l_{a}+1} d_{k}^{2}\left(\mathcal{D}_{s}^{l}, \mathcal{D}_{t}^{l}\right)
  \label{loss_fuc}
\end{equation}
where $\mathcal{L}_{C}\left(X_{s},Y_{s}\right)$ is a CD loss on the source labeled data, $l_{a}$ is layer index, $d_{k}^{2}\left(\mathcal{D}_{s}^{l}, \mathcal{D}_{t}^{l}\right)$ indicates the MK-MMD between the two domains on the difference features of the $l_{a}$-th layer, and $\lambda>0$ denotes a domain adaptation penalty factor. 

\subsubsection{Optimization}
\par In the training procedure of DSDANet, two types of parameters require to be learned, one is the network parameters $\Theta$ and another is the kernel coefficient $\beta$. However, the cost of MK-MMD computation by kernel trick is $O\left(n^{2}\right)$, it is unacceptable for deep models, especially in large-scale CD data sets and makes the training procedure more difficult. Therefore, the unbiased estimate of MK-MMD is utilized to decrease the computation cost from $O\left(n^{2}\right)$ to $O\left(n\right)$ \cite{Gretton2012}, which can be formulated as
\begin{equation}
  \left\{\begin{aligned}
    & d_{k}^{2}\left(D_{s}^{l}, D_{t}^{l}\right)=\frac{2}{n_{s}} \sum\limits_{i=1}^{n_{s} / 2} g_{k}\left(z_{i}^{l}\right) \\ 
    & g_{k}\left(z_{i}\right)=k\left(h_{2 i-1}^{s l}, h_{2 i}^{s l}\right)+k\left(h_{2 i-1}^{t l}, h_{2 i}^{t l}\right) \\ 
    & \quad \quad \ \ \  -k\left(h_{2 i-1}^{s l}, h_{2 i}^{t l}\right)-k\left(h_{2 i}^{s l}, h_{2 i-1}^{t l}\right)
    \end{aligned}\right.
\end{equation}
where $z_{i}^{l}=\left(h_{2 i-1}^{s l}, h_{2 i}^{s l}, h_{2 i-1}^{t l}, h_{2 i}^{t l}\right)$ is a quad-tuple evaluated by multi-kernel $k$ and $h^{l}$ is learned features in $l$-th layer. Thus, the deviation of $\mathcal{L}$ to the parameters of the adaptation layer $\Theta^{l}$ can be expressed as
\begin{equation}
  \frac{\partial \mathcal{L}}{\partial \Theta^{l}}=\frac{\partial \mathcal{L}_{C}}{\partial \Theta^{l}}+\lambda \frac{\partial g_{k}\left(z_{i}^{l}\right)}{\partial \Theta^{l}}
\end{equation}

\par As for the kernel parameters $\beta$, the optimal coefficient for each $d_{k}^{2}\left(D_{s}^{l}, D_{t}^{l}\right)$ can be sought by jointly maximizing $d_{k}^{2}\left(D_{s}^{l}, D_{t}^{l}\right)$ itself and minimizing the variance \cite{Gretton2012}, which results in the optimization
\begin{equation}
  \max _{k \in \mathcal{K}} d_{k}^{2}\left(\mathcal{D}_{s}^{\prime}, \mathcal{D}_{t}^{\prime}\right) \sigma_{k}^{-2}
  \label{beta_opt}
\end{equation}
where $\sigma_{k}^{2}=E\left(g_{k}^{2}(z)\right)-E^{2}\left(g_{k}(z)\right)$ is estimation variance of $g_{k}$. Further, denoting covariance of $g_{k}$ as $Q=\operatorname{cov}\left(g_{k}\right) \in \mathcal{R}^{n \times n}$, each element of $Q$ can be calculated as $Q_{u, v}=\frac{4}{n_{s}} \sum_{i=1}^{4 / n_{s}} g_{\Delta, k_{u}}\left(\bar{z}_{i}\right) g_{\Delta, k_{v}}\left(\bar{z}_{i}\right)$, $g_{\Delta, k_{u}}\left(\bar{z}_{i}\right)=g_{k_{u}}\left(z_{2 i-1}\right)-g_{k_{u}}\left(z_{2 i}\right)$. Eventually, the optimization (\ref{beta_opt}) can be resolved as a quadratic program
\begin{equation}
  \begin{array}{l}
    \min \beta^{T} Q \beta \\
    \text {s.t.}\left\{\begin{array}{l}
    d^{T} \beta=1 \\
    \beta \geq 0
    \end{array}\right.
    \end{array}
  \label{QP_pro}
\end{equation}
where $d=\left[d_{1}, d_{2}, \ldots, d_{n}\right]^{T}$, $d_{i}$ is the MMD between two domains based on kernel $k_{i}$. 

\par By alternatively adopting optimizer, such as stochastic gradient descent (SGD), to update $\Theta$ and solving the quadratic program problem (\ref{QP_pro}) to optimize $\beta$, the DSDANet can gradually learn transferrable difference representation from source labeled data and target unlabeled data. By minimizing Eq. (\ref{loss_fuc}), the marginal distributions $P\left(X_{s}\right)$ and $P\left(X_{t}\right)$ of two domains would become very similar, yet the conditional distributions $P\left(Y_{s} | X_{s}\right)$ and $P\left(Y_{t} | X_{t}\right)$ of two domains may still be slightly different. Thus, a very small part of target labeled data is selected to fine-tune DSDANet. Compared with the enough labeled data in the source domain, the labeled data provided by the target domain is very limited, so this procedure can be treated as a semi-supervised learning fashion. 

\begin{figure}[!t]

  \centering
  \includegraphics[scale=0.28]{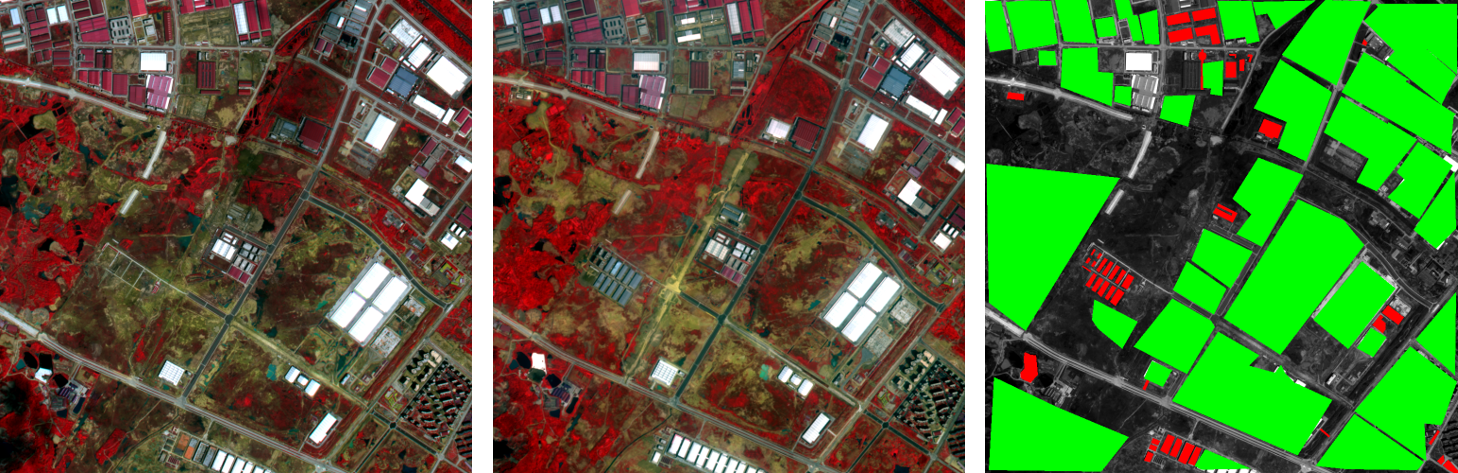}
  \caption{The WH data set used as source domain. Left: pre-change. Middle: post-change. Right: ground truth, where red is change and green indicates non-change.}
  \label{fig:SDDS}
\end{figure}

\begin{figure}[ht]
  \centering

  \subfloat[]{
    \includegraphics[scale=0.28]{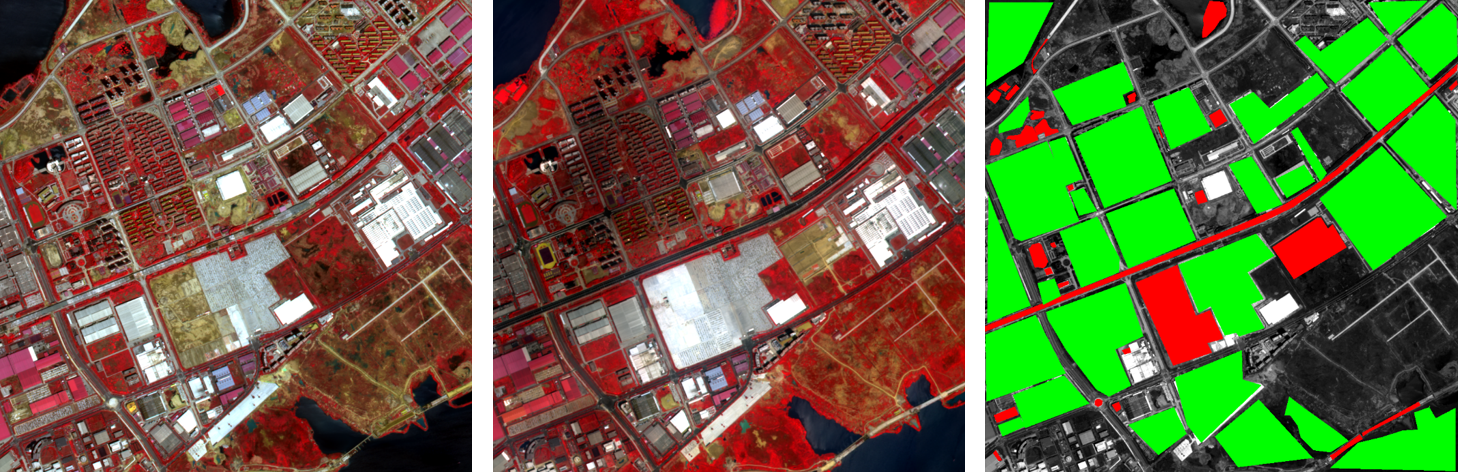}
  \label{fig_first_case}}
  \hfil
  \subfloat[]{
    \includegraphics[scale=0.28]{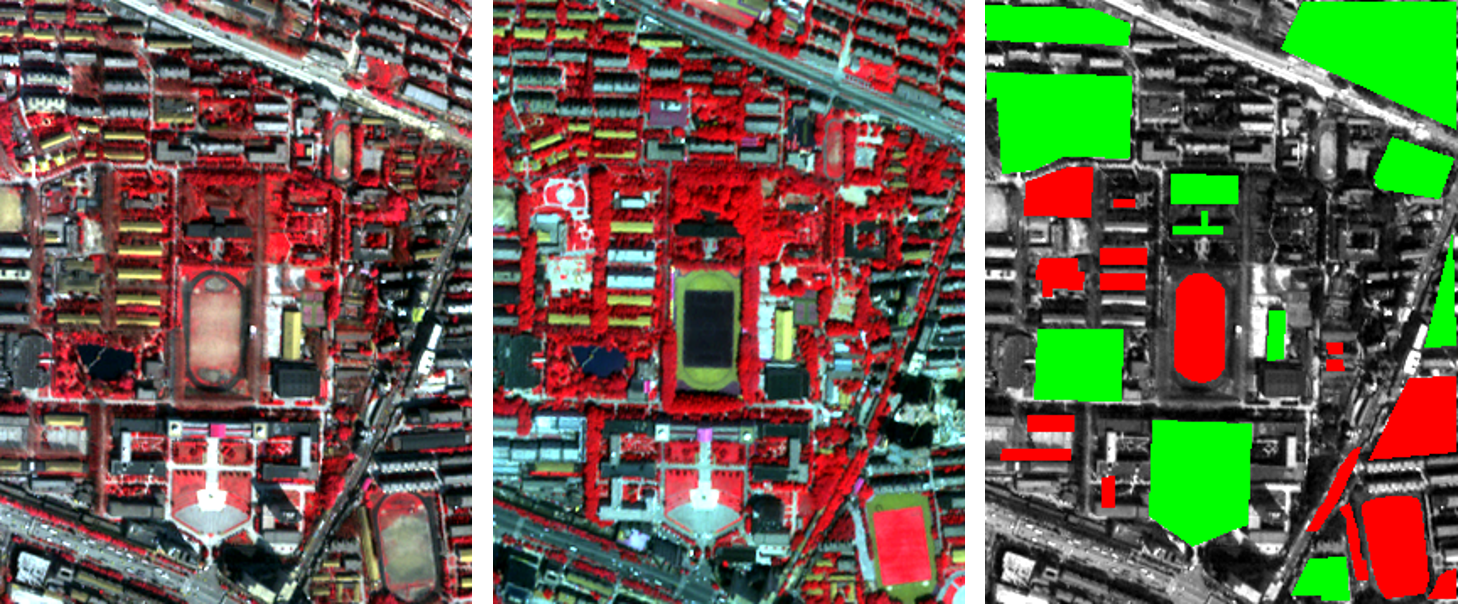}
  \label{fig_second_case}}
  \hfil
  \subfloat[]{
    \includegraphics[scale=0.28]{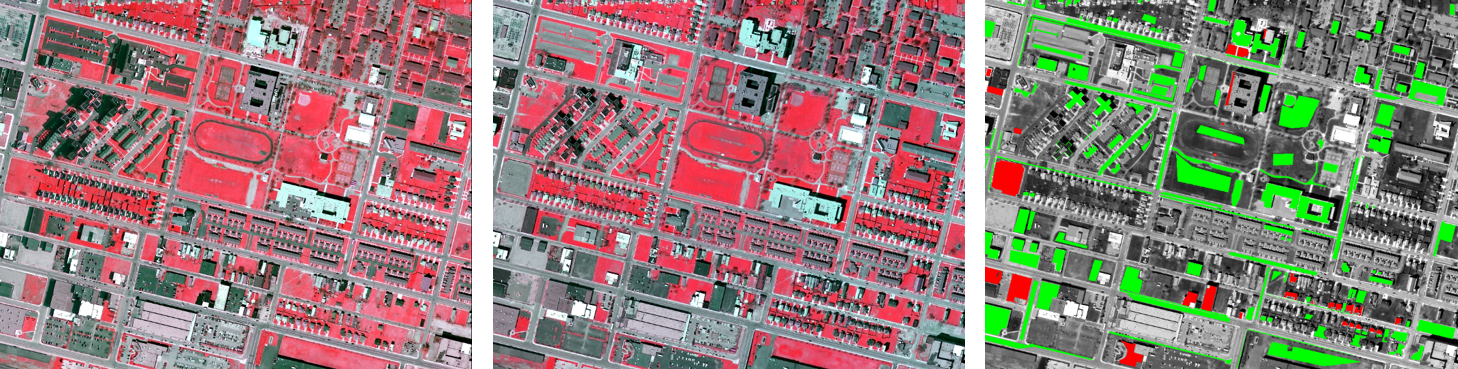}
  \label{fig_third_case}}

  \caption{Three data sets used as target domains. (a) HY data set. (b) QU data set. (c) LO data set. Left: pre-change. Middle: post-change. Right: ground truth, where red is change and green indicates non-change.}
  \label{fig:TDDS}

\end{figure}

\section{Experiment}\label{sec:4}
\subsection{Data Description}
\par To validate the effectiveness of our proposed in cross-domain CD, experiments are performed on four CD data sets. Among the four data sets, WH data set, as shown in Fig. \ref{fig:SDDS}, is adopted as source domain. The two multi-temporal images were captured by GF-2 on April 4, 2016 and September 1, 2016, covering the city of Wuhan, China. The size of the two images is 1000$\times$1000 pixels with a spatial resolution of 4m and four spectral bands. The main changes between the two images are groundwork before building over, new built-up regions. 

\par The three data sets adopted as target domains are shown in Fig. \ref{fig:TDDS}. The first is HY data set, it is also captured by GF-2, covering the city of Hanyang, China. The size of the two images is 1000$\times$1000 pixels with a spatial resolution of 4m and four spectral bands. But compared with WH data set, the scene of HY data set is more complicated and the types of change are more abundant, mainly involving building, vegetation, bare soil, and water. 

\par The second target data set is QU data set, illustrated in Fig. \ref{fig:TDDS}-(b), which was acquired by QuickBird in 2002 and 2006 with a spatial resolution of 2.4m and four spectral bands. Both images in this data set are 358$\times$280 pixels. The area covered by QU data set is Wuhan University. Ground truth shows the main changes during the four years, including buildings and some infrastructures. 

\par The last target data set (shown in Fig. \ref{fig:TDDS}-(c)) used in our experiments, called LO, is provided by the New York State Digital Orthoimagery Program. The two images acquired in 2008 and 2011 contain four bands with a size of 800$\times$1024 pixels and a resampled resolution of 1m. This data set covers a part of the downtown area in Buffalo, NY, USA. The detailed information of four data sets is also listed in Table \ref{DS_Info}.
											
\begin{table*}[t]
  \centering
  \renewcommand{\arraystretch}{1.4}
  \caption{INFORMATION OF THE FOUR DATA SETS}
  \label{DS_Info}
  \begin{tabular}{c c c c c c c}
    \hline
    \bfseries Data set & \bfseries Sensor & \bfseries Size & \bfseries Area & \bfseries T1 & \bfseries T2 & \bfseries Resolution \\
    \hline\hline
    WH	& GF-2	& 1000$\times$1000	& Wuhan, China	& 2016	& 2016 & 4m \\
    HY	& GF-2	& 1000$\times$1000	& Hanyang, China	& 2016	& 2016 & 4m \\				
    QU	& QuickBird	& 358$\times$280	& Wuhan, China	& 2002	& 2006 & 2.4m \\					
    LO	& Airborne & 800$\times$1024	& Buffalo, USA	& 2008	& 2011 & 1m \\					
    \hline
  \end{tabular}
\end{table*}

\begin{table*}[t]
  \centering
  \renewcommand{\arraystretch}{1.4}
  \caption{SPECIFIC NETWORK STRUCTURE AND CONFIGURATION OF DSDANET}
  \label{Net_Info}
  \begin{tabular}{ccccccc}
    \hline
    \bfseries Layer & \bfseries Input Shape & \bfseries Output Shape & \bfseries Number of Filters & \bfseries Configuration \\
    \hline\hline
    Conv1-1	& (8, 8, 4)	& (8, 8, 16)	& 16	& 3$\times$3 kernel, padding 1, stride 1, ReLU \\
    Conv1-2	& (8, 8, 16)	& (8, 8, 16)	& 16	& 3$\times$3 kernel, padding 1, stride 1, ReLU \\
    Max-pooling1	& (8, 8, 16)	& (4, 4, 16)	& -	& 2$\times$2 kernel, stride 2 \\
    Conv2-1	& (4, 4, 16)	& (4, 4, 32)	& 32	& 3$\times$3 kernel, padding 1, stride 1, ReLU \\
    Conv2-2	& (4, 4, 32)	& (4, 4, 32)	& 32	& 3$\times$3 kernel, padding 1, stride 1, ReLU  \\
    Max-pooling2	& (4, 4, 32)	& (2, 2, 32)	& 16	& 2$\times$2 kernel, stride 2  \\	
    Conv3-1	& (2, 2, 32)	& (2, 2, 64)	& 64	& 3$\times$3 kernel, padding 1, stride 1, ReLU \\
    Conv3-2	& (2, 2, 64)	& (2, 2, 64)	& 64	& 3$\times$3 kernel, padding 1, stride 1, ReLU  \\
    Max-pooling3	& (2, 2, 64)	& (1, 1, 64)	& 64	& 2$\times$2 kernel, stride 2  \\				
    \hline
    Difference	& (1, 1, 64)	& (64,)	& -	& -  \\		
    FC-1	& (64,)	& (128,)	& -	& embedded with MK-MMD, ReLU\\		
    FC-2	& (128,)	& (128,)	& -	& embedded with MK-MMD, ReLU  \\		
    FC-3	& (128,)	& (1,)	& -	& Sigmoid \\		
    \hline
  \end{tabular}
\end{table*}

\subsection{Experiment Settings}
\par Based on the source data set and three target data sets, we design three representative cross-domain CD tasks, i.e., WH$\rightarrow$HY, WH$\rightarrow$QU and WH$\rightarrow$LO, and evaluate our method on the three tasks. For WH$\rightarrow$HY, both domains were captured by the same sensor but with different imaging conditions and covered area, which is the case 1 of cross-domain CD. For the latter two tasks, since the source domain and the target domain were acquired by different sensors leading to diverse spatial resolutions, covering area and statistical characteristics, the data distributions of two domains are significantly different, which belongs to the case 2 of cross-domain CD. 

\par In the three cross-domain CD tasks, the specific network structure of DSDANet is consistent and listed in Table \ref{Net_Info}. Note that the two convolutional streams of DSDANet are siamese and have the same structure, thus we just list the structure of one branch in Table \ref{Net_Info}. In the training procedure of each task, we randomly select 5000 samples from the source domain as labeled training samples. And we train DSDANet with labeled source training samples and all target samples without labels. After training, we only select 20 labeled samples (10 changed samples, 10 unchanged samples) from the target domain for fine-tuning. Compared with the labeled source data, the labeled data provided by the target domain cannot cover all types of changes and non-changes, it is highly sparse and is easy to collect in practice. For the training procedure, Adam optimizer \cite{Kingma2014} with a learning rate of 1$e^{-3}$ is adopted to minimize the CD loss and domain adaptation discrepancy simultaneously. The optimizer for fine-tuning is SGD with a learning rate of 1$e^{-4}$ and a momentum of 0.9.

\par DSDANet uses MK-MMD to bridge the discrepancy between two domains, in this paper, a family of $n$ Gaussian kernels is applied, the bandwidth $\gamma_{n}$ are varying from 2$e^{-7}\gamma$ to 2$e^{7}\gamma$ with a multiplication step of 2, where $\gamma$ is the median pairwise distances between data of two domains \cite{Gretton2012}. For domain penalty factor $\lambda$, we set 0.2, 0.8, and 0.6 for WH$\rightarrow$HY, WH$\rightarrow$QU, WH$\rightarrow$LO, respectively. 

\par To validate the superiority of the proposed model, six commonly used CD models and three variants of DSDANet are chosen as comparison methods, which are listed as follows.
\begin{enumerate} 
  \item IRMAD \cite{Nielsen2007}, which is an iteratively weighted extension of MAD and has shown good performance in multispectral image CD.
  \item CVA \cite{Sharma2007}, which is one of the most classic unsupervised CD approaches and is effective for CD in multispectral images.
  \item OBCD \cite{Desclee2006}, an effective unsupervised CD method that adopts the object generated by segmentation algorithms as the basic unit of CD.
  \item SVM, a supervised machine learning model, has shown success in many fields, which can construct hyperplane in high-dimensional space to separate change and non-change. 
  \item RNN \cite{Lyu2016}, a deep learning-based method that has been proven to learn transferable rules for land-cover CD. 
  \item DSFANet \cite{Du2019a}, a deep learning-based CD model, which works by extracting deep features from multispectral images with two-streams DNN and detecting changes with SFA. 
  \item DSCNet, the backbone of our DSDANet, which has the same network structure with DSDANet, but not utilizing domain adaptation to minimize the distance between two domains. 
  \item DSDANet-SK, a variant of our DSDANet that uses MMD with a single kernel in adaptation layers for domain adaptation. 
  \item DSDANet-SL, a variant of our DSDANet that only embeds MK-MMD in the penultimate FC layer instead of multiple layers.
  \item DSDANet, our proposed model that restricts domain discrepancy by MK-MMD with multi-layer adaptation schema. 
\end{enumerate}

\begin{figure*}[t]
  \centering
  \subfloat[]{
    \includegraphics[width=1.25in]{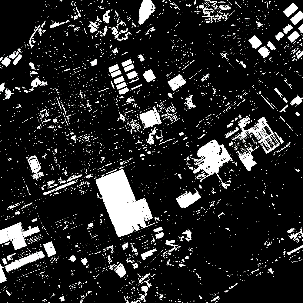}
  \label{HY_IRMAD}}
  \hfil
  \subfloat[]{
    \includegraphics[width=1.25in]{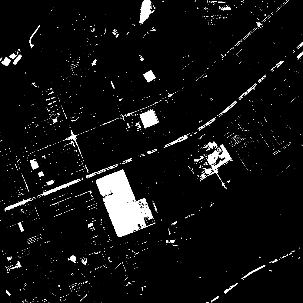}
  \label{HY_CVA}}
  \hfil
  \subfloat[]{
    \includegraphics[width=1.25in]{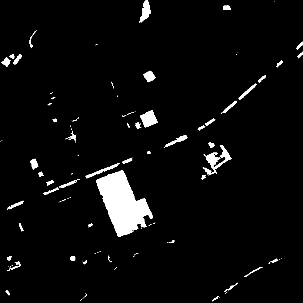}
  \label{HY_OBCD}}
  \hfil
  \subfloat[]{
    \includegraphics[width=1.25in]{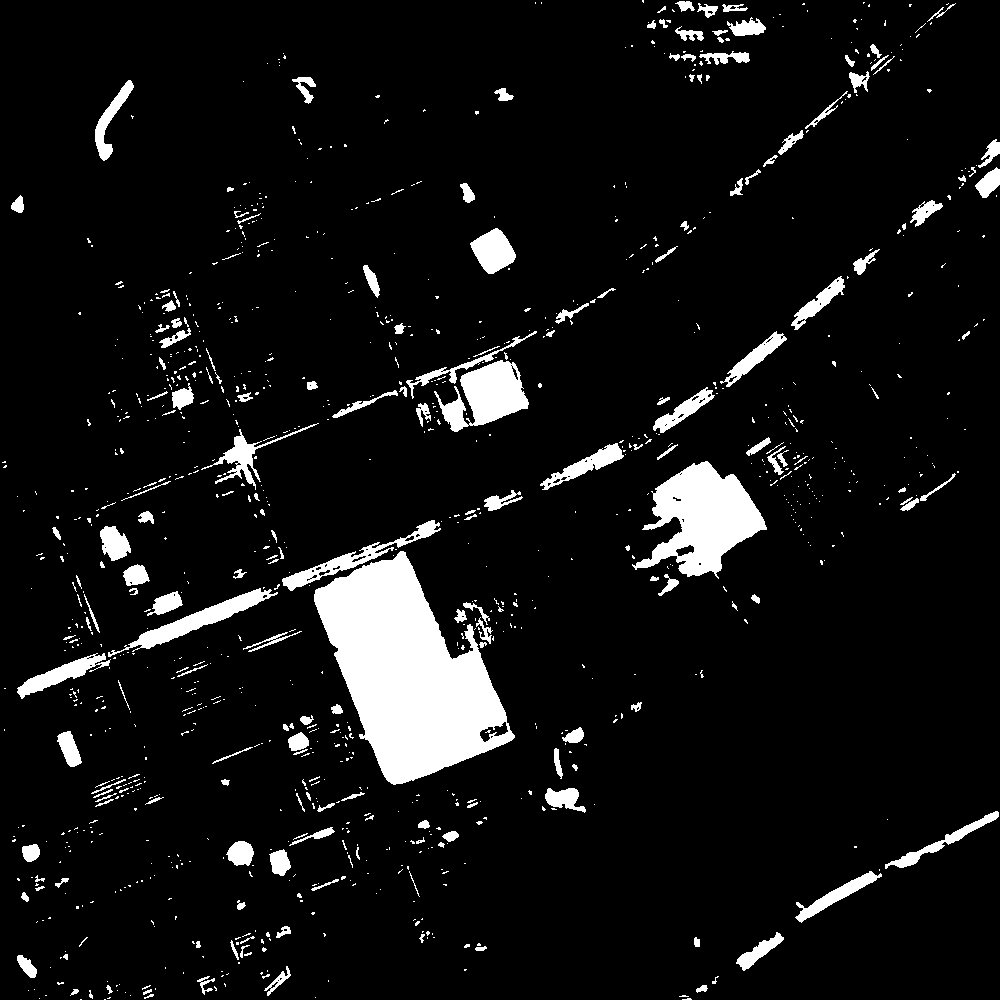}
  \label{HY_SVM}}
  \hfil
  \subfloat[]{
    \includegraphics[width=1.25in]{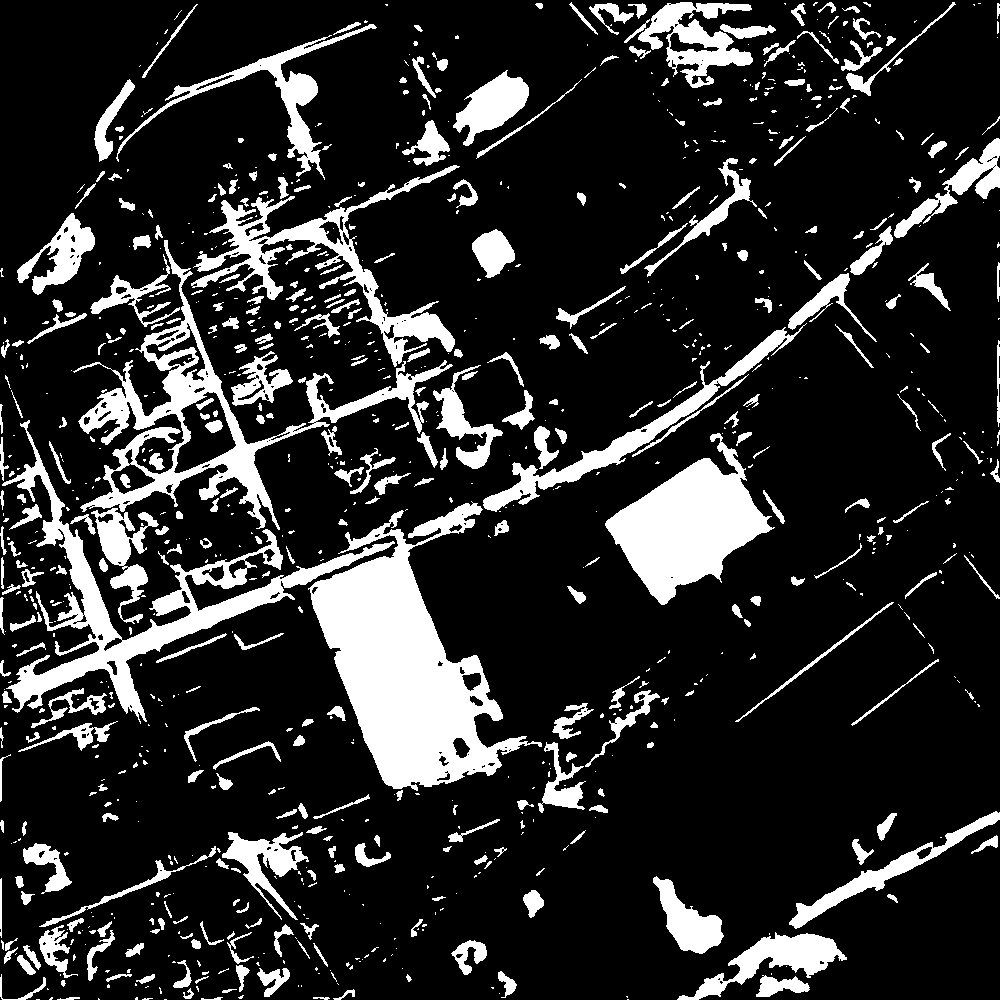}
  \label{HY_RNN}}

  \subfloat[]{
    \includegraphics[width=1.25in]{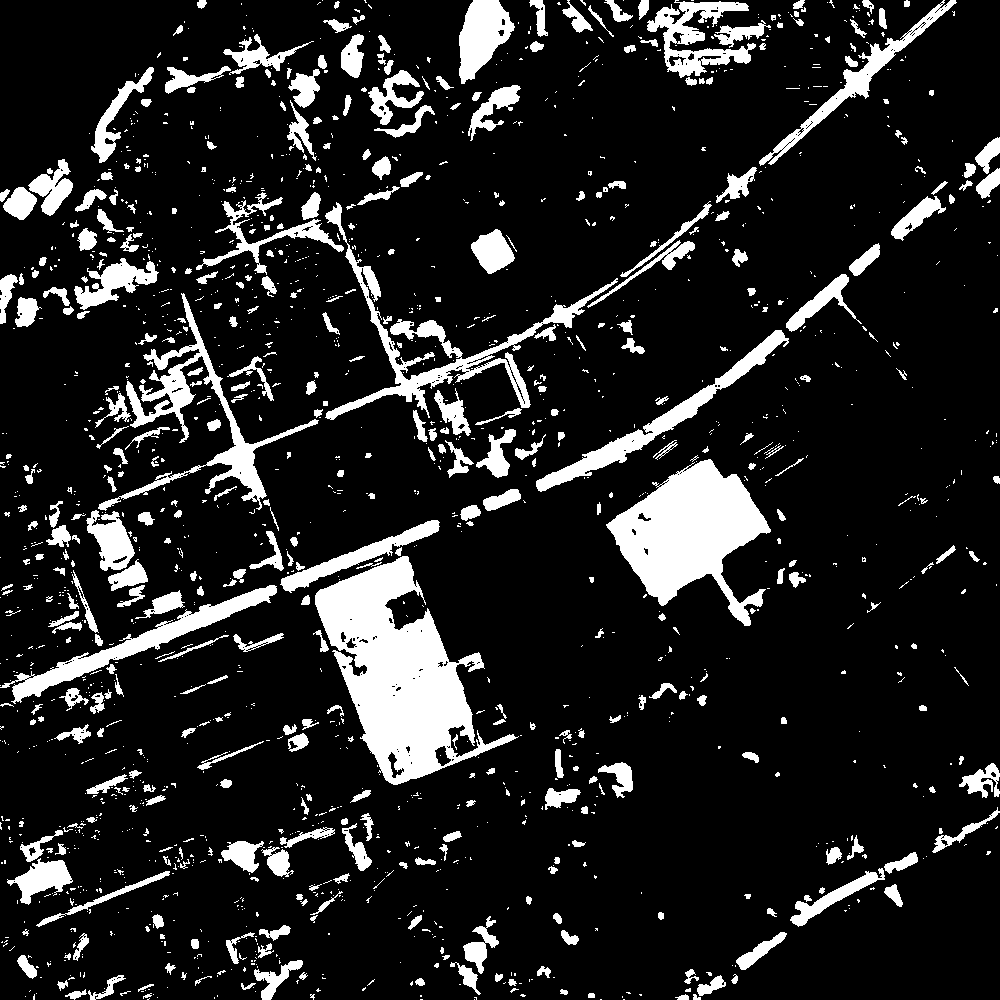}
  \label{HY_DSFANet}}
  \hfil
  \subfloat[]{
    \includegraphics[width=1.25in]{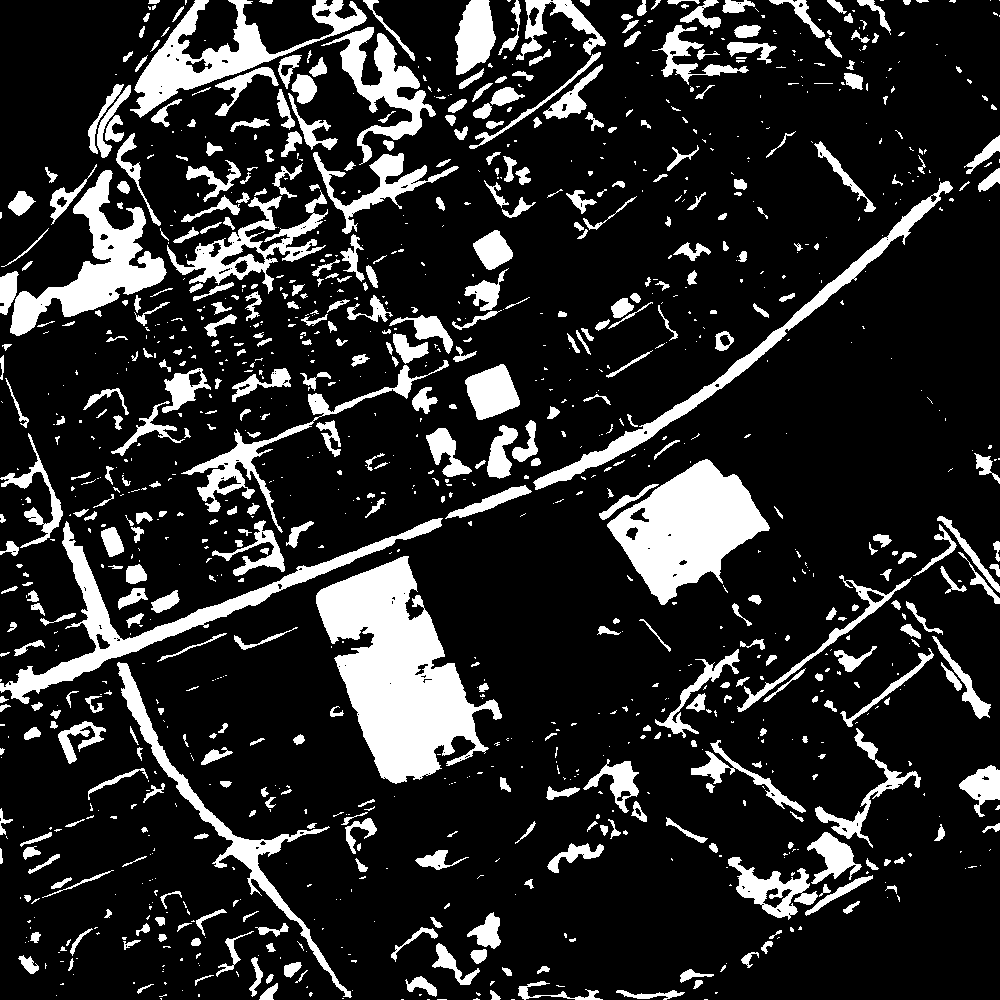}
  \label{HY_DSCNet}}
  \hfil
  \subfloat[]{
    \includegraphics[width=1.25in]{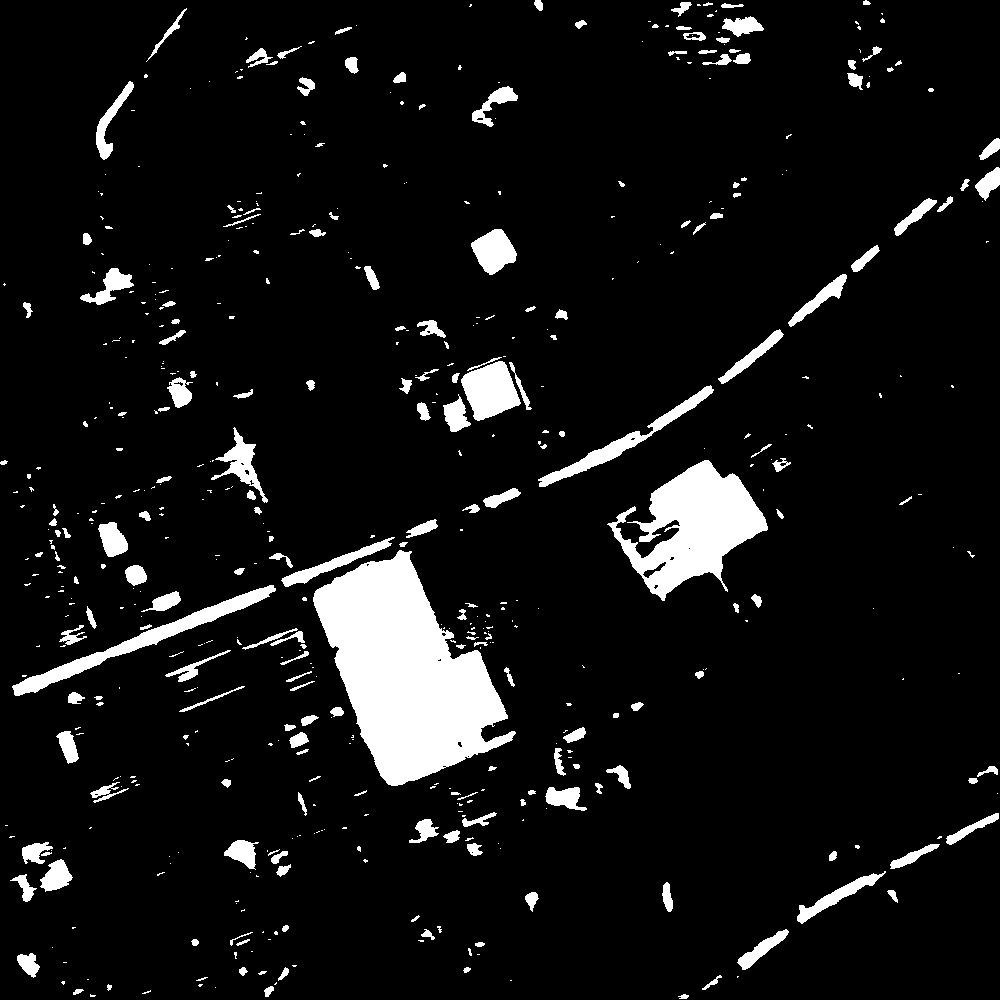}
  \label{HY_DSDANet_SK}}
  \hfil
  \subfloat[]{
    \includegraphics[width=1.25in]{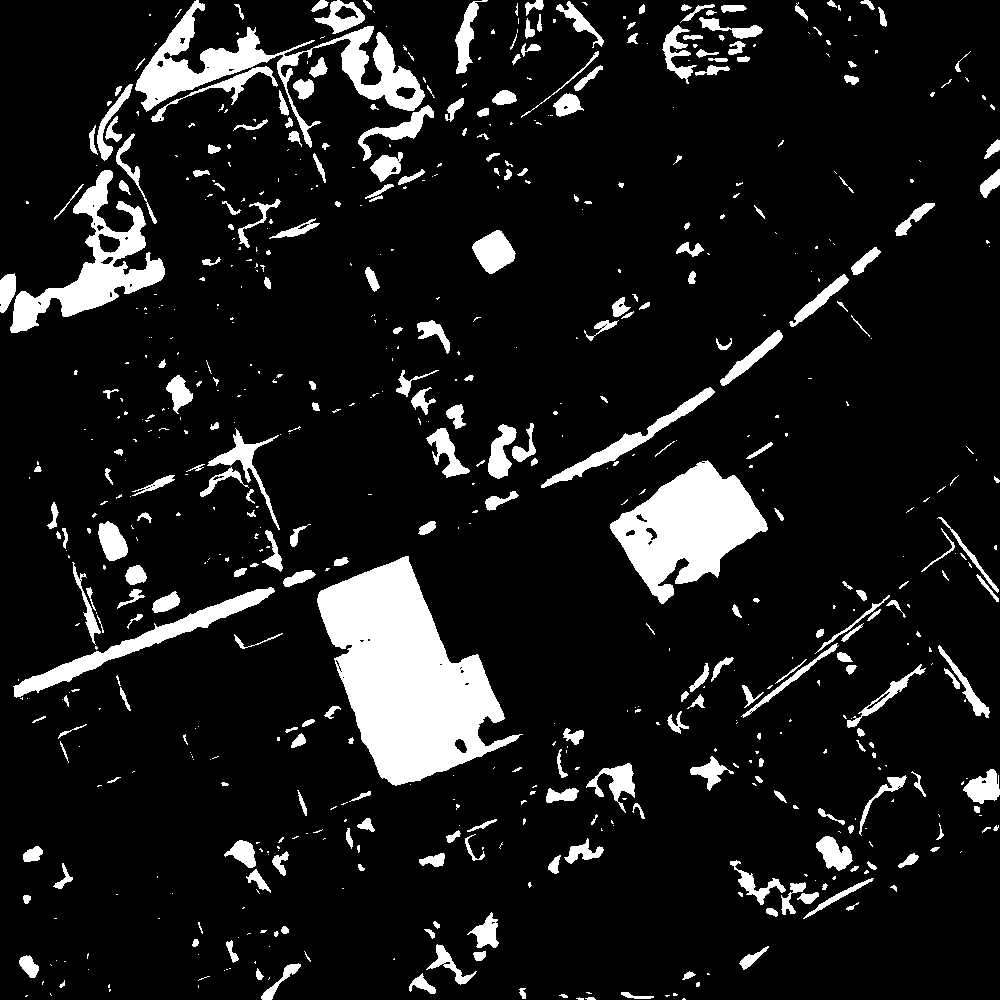}
  \label{HY_DSDANet_SL}}
  \hfil
  \subfloat[]{
    \includegraphics[width=1.25in]{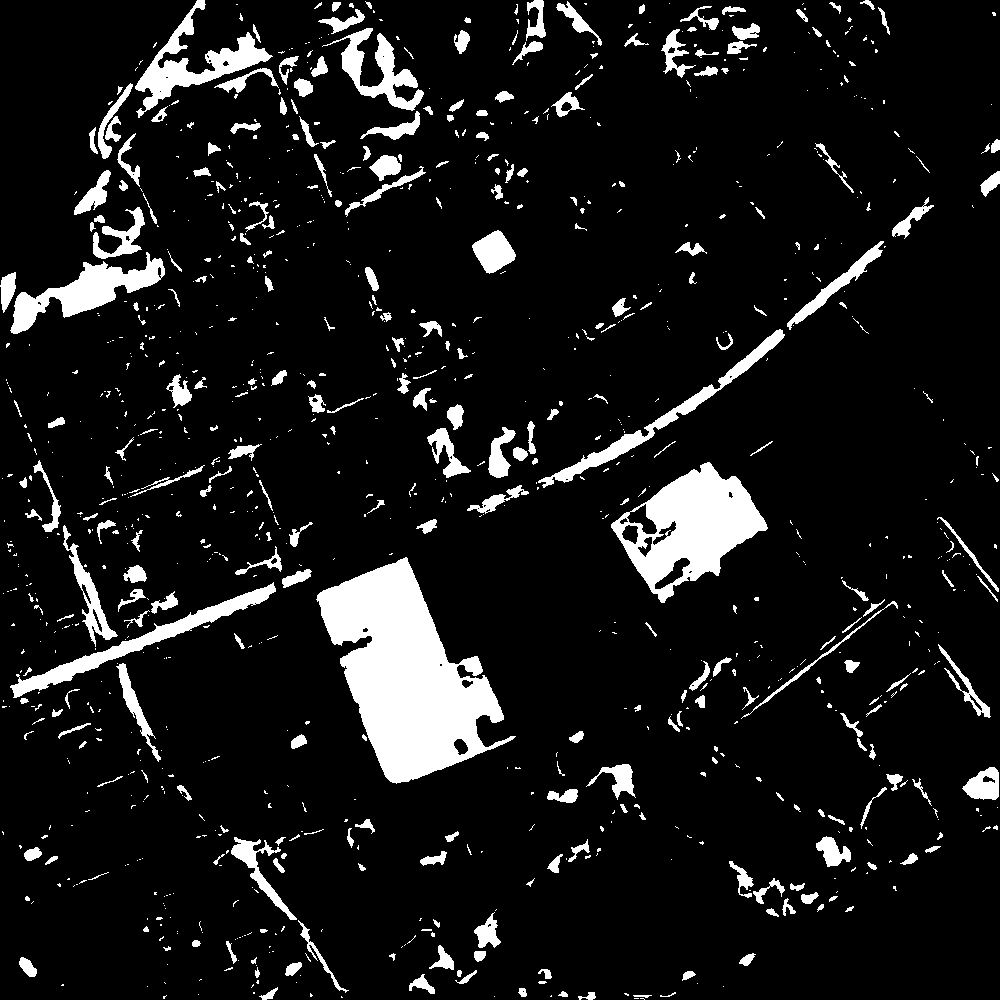}
  \label{HY_DSDANet}}

  \caption{Binary change maps obtained by different methods on the HY data set. (a) IRMAD. (b) CVA. (c) OBCD. (d) SVM. (e) RNN. (f) DSFANet. (g) DSCNet. (h) DSDANet-SK. (i) DSDANet-SL. (j) DSDANet. }
  \label{HY_result}
\end{figure*}

\par Among these methods, IRMAD, CVA, and OBCD are unsupervised models, they are directly performed on three target data sets to get CD results. DSFANet is a semi-supervised model requiring unchanged labeled data, we train it on the target labeled data randomly selected from the ground truth. The other five models are all supervised, for SVM, we train it on both source labeled data and sparse target labeled data. Radial basis function (RBF) is adopted as the kernel function of SVM and the optimal hyper-parameters $C$ (penalty factor) and $\gamma$ (bandwidth of RBF) are traced by grid search. For the remaining four deep learning-based methods, we train them on the source labeled data and fine-tune them with sparse target labeled data. 

\par Lastly, to quantitatively evaluate the performance of different methods from various aspects, five evaluation criteria are used for accuracy assessment, including false positive (FP), false negative (FN), overall error (OE), overall accuracy (OA) and kappa coefficient (KC). FP and FN denote the number of pixels that are falsely detected as changed pixels and unchanged ones, respectively. OE represents the total number of misclassified pixels, OE=FP+FN. OA shows the number of correctly detected pixels, divided by the number of total pixels, can be calculated as OA=1-OE / N, where N is the number of total pixels. KP is a comprehensive index measuring the similarity between the experimental results and ground truth, which is defined as KP=(OA-PE) / (1-PE), where PE=((TP+FP) (TP+FN) +(TN+FN) (FP+TN)) / $N^{2}$.

\begin{table}[t]
  \footnotesize 
  \centering
  \renewcommand{\arraystretch}{1.3}
  \caption{ACCURACY ASSESSMENT ON CHANGE DETECTION RESULTS OBTAINED BY DIFFERENT METHODS ON HY DATA SET.}
  \label{HY_table}
  \begin{tabular}{c c c c c c}
    \hline
    \bfseries Method & \bfseries FP & \bfseries FN & \bfseries OE & \bfseries OA & \bfseries KC\\
    \hline\hline
    IRMAD	& 54408	& 17063	& 71471	& 84.97	& 0.4565 \\
    CVA	& 6483	& 19896	& 26379	& 94.45	& 0.7174 \\ 					
    OBCD	& \textbf{775}	& 22585	& 23360	& 95.08	& 0.7310 \\ 					
    SVM & 18619	& \textbf{6737}	& 25356	& 94.67	&	0.7311 \\ 					
    RNN	& 40328	& 11208	& 51533	& 89.16	& 0.5882 \\ 				
    DSFANet	& 17727	& \underline{8147}	& 25784	& 94.56	& 0.7657 \\ 
    DSCNet	& 23571	& 10675	& 34246	& 92.80	& 0.6967 \\ 		
    \hline			
    DSDANet-SK	& 7751	& 14299	& 22050	& 95.36	& 0.7757 \\ 		
    DSDANet-SL	& \underline{5552}	& 13061	& \textbf{18613}	& \textbf{96.08}	& \textbf{0.8093} \\
    DSDANet	& 5787	& 13921	& \underline{19708}	& \underline{95.85}	& \underline{0.7971} \\ 					
    \hline
  \end{tabular}
\end{table}

\subsection{Experimental Results and Analysis}
\par The CD results obtained by different methods on the three source domains are shown in Fig. \ref{HY_result}-Fig. \ref{LO_result} and the values of evaluation criteria are listed in Table \ref{HY_table}-Table \ref{LO_table}. The following conclusions can be drawn according to the experimental results:

\subsubsection{Domain adaptation vs standard fine-tuning}
\par
\begin{figure*}[!t]
  \centering
  \subfloat[]{
    \includegraphics[width=1.25in]{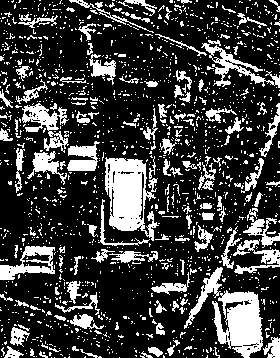}
  \label{QU_IRMAD}}
  \hfil
  \subfloat[]{
    \includegraphics[width=1.25in]{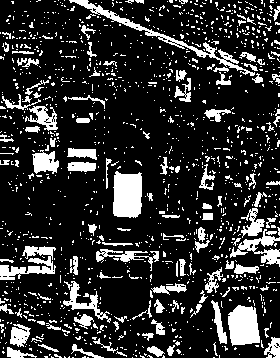}
  \label{QU_CVA}}
  \hfil
  \subfloat[]{
    \includegraphics[width=1.25in]{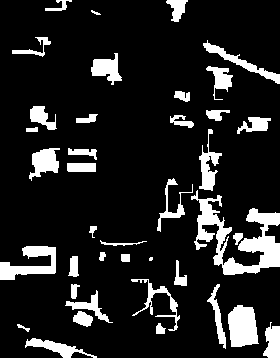}
  \label{QU_OBCD}}
  \hfil
  \subfloat[]{
    \includegraphics[width=1.25in]{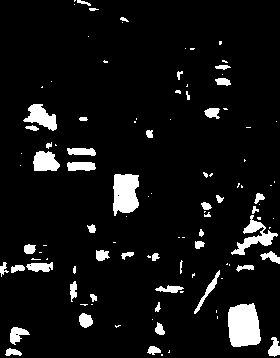}
  \label{QU_SVM}}
  \hfil
  \subfloat[]{
    \includegraphics[width=1.25in]{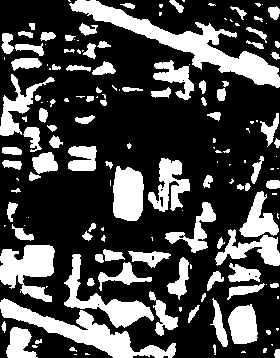}
  \label{QU_RNN}}

  \subfloat[]{
    \includegraphics[width=1.25in]{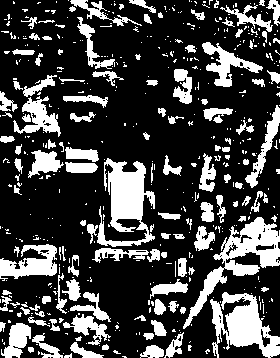}
  \label{QU_DSFANet}}
  \hfil
  \subfloat[]{
    \includegraphics[width=1.25in]{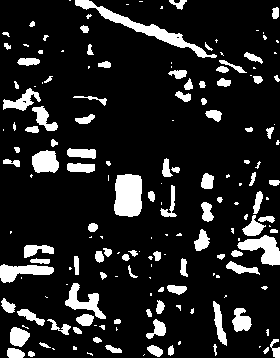}
  \label{QU_DSCNet}}
  \hfil
  \subfloat[]{
    \includegraphics[width=1.25in]{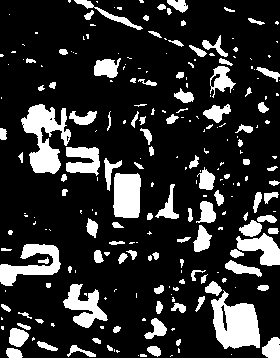}
  \label{QU_DSDANet_SK}}
  \hfil
  \subfloat[]{
    \includegraphics[width=1.25in]{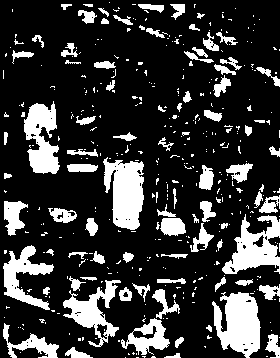}
  \label{QU_DSDANet_SL}}
  \hfil
  \subfloat[]{
    \includegraphics[width=1.25in]{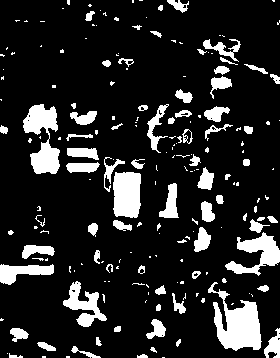}
  \label{QU_DSDANet}}

  \caption{Binary change maps obtained by different methods on the QU data set. (a) IRMAD. (b) CVA. (c) OBCD. (d) SVM. (e) RNN. (f) DSFANet. (g) DSCNet. (h) DSDANet-SK. (i) DSDANet-SL. (j) DSDANet. }
  \label{QU_result}
\end{figure*}

\begin{figure*}[t]
  \centering
  \subfloat[]{
    \includegraphics[width=1.25in]{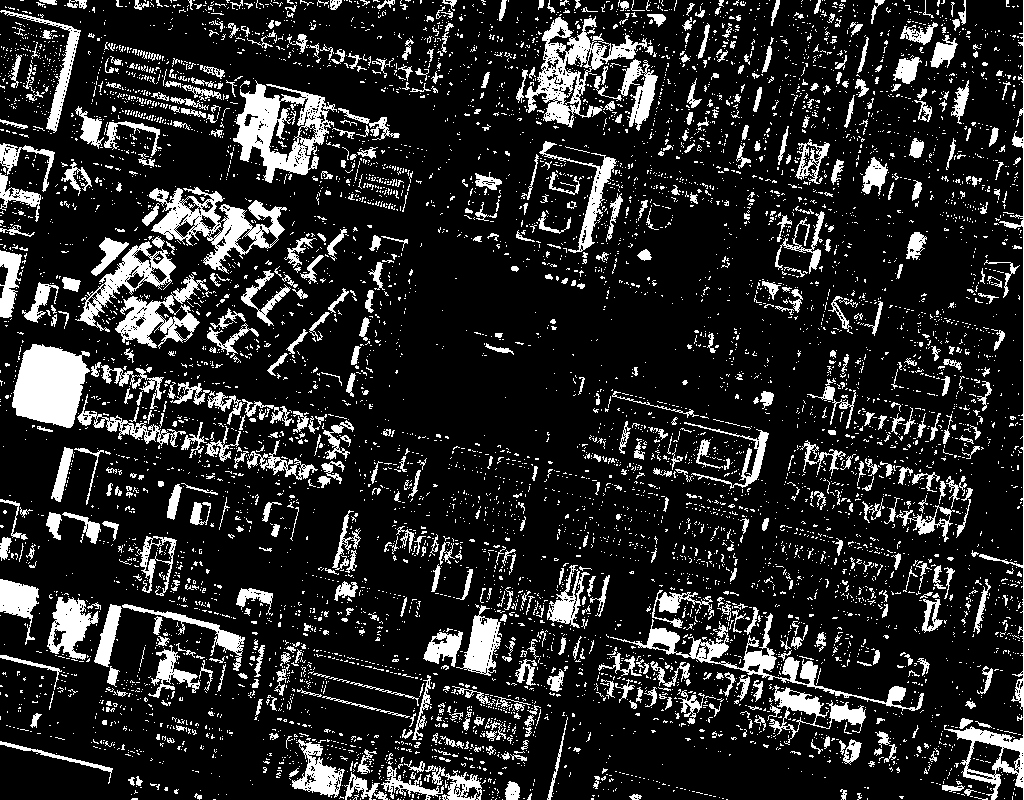}
  \label{LO_IRMAD}}
  \hfil
  \subfloat[]{
    \includegraphics[width=1.25in]{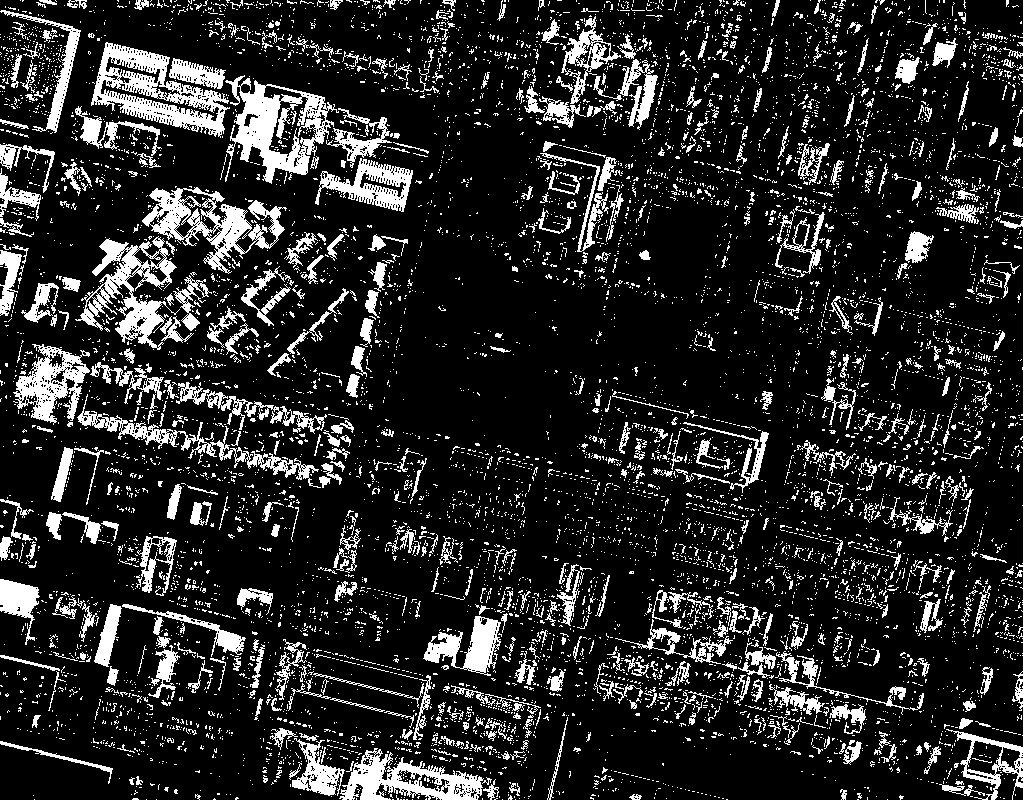}
  \label{LO_CVA}}
  \hfil
  \subfloat[]{
    \includegraphics[width=1.25in]{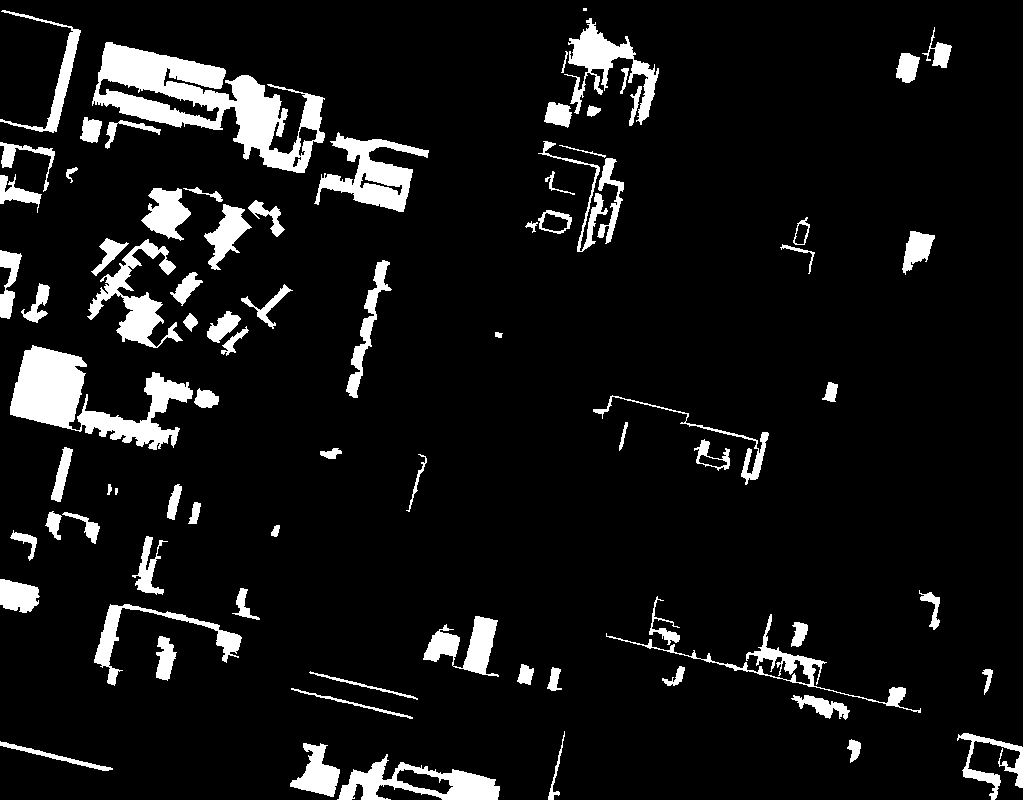}
  \label{LO_OBCD}}
  \hfil
  \subfloat[]{
    \includegraphics[width=1.25in]{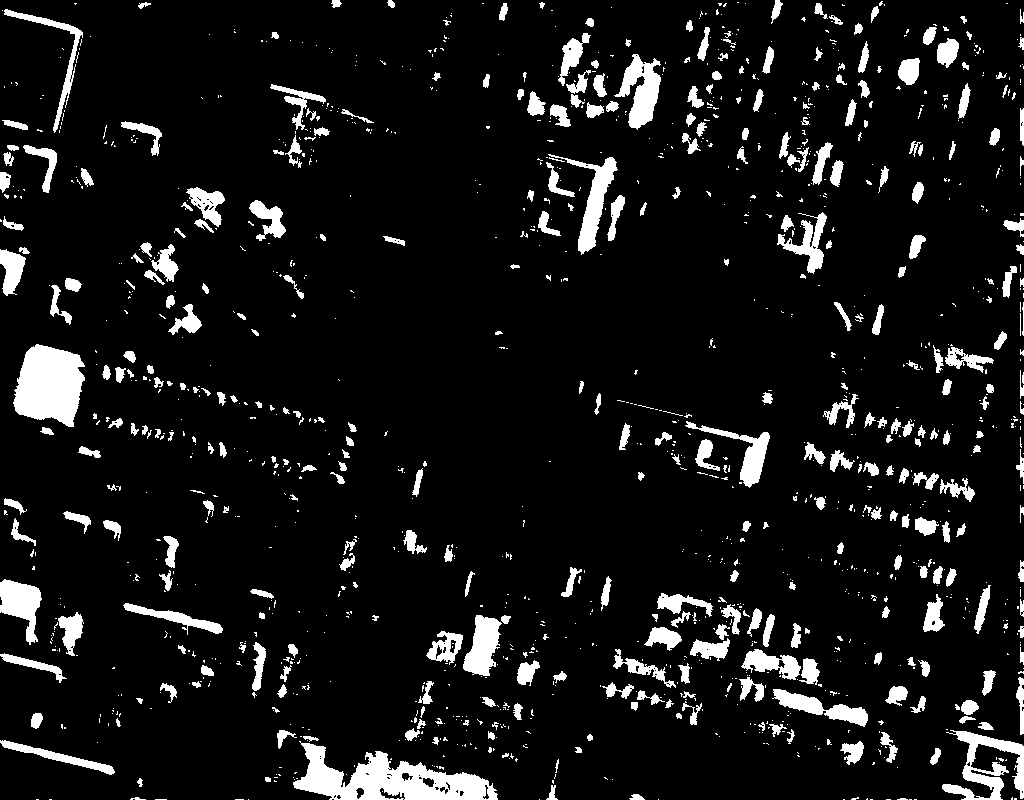}
  \label{LO_SVM}}
  \hfil
  \subfloat[]{
    \includegraphics[width=1.25in]{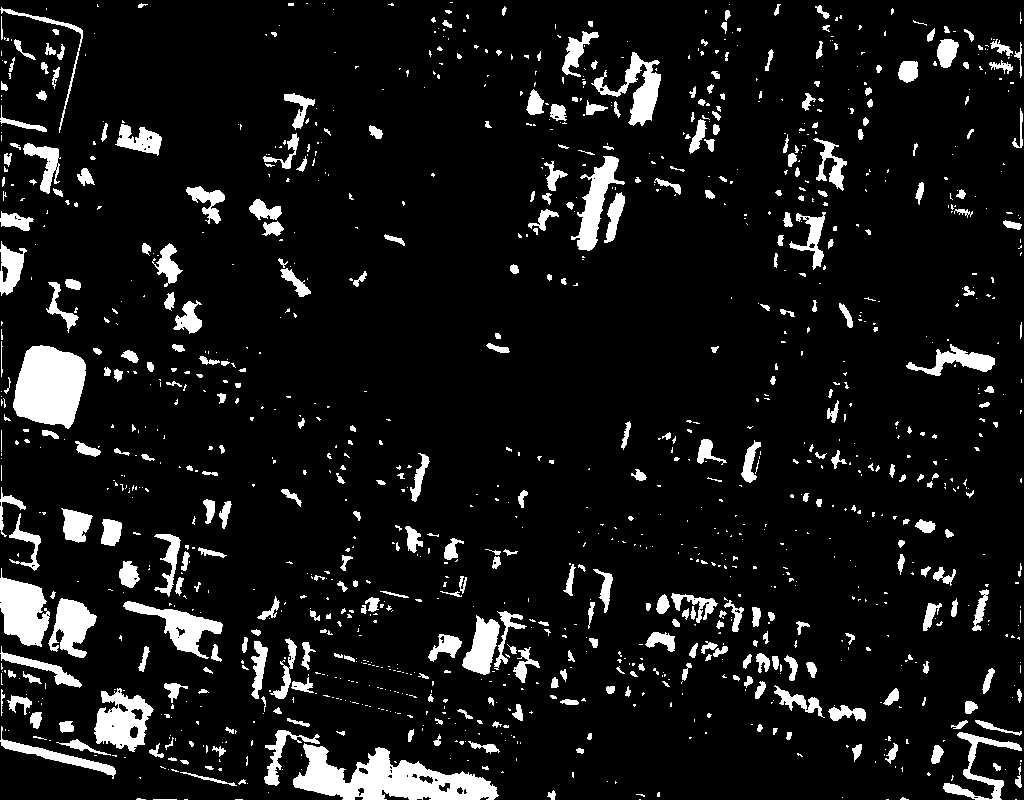}
  \label{LO_RNN}}

  \subfloat[]{
    \includegraphics[width=1.25in]{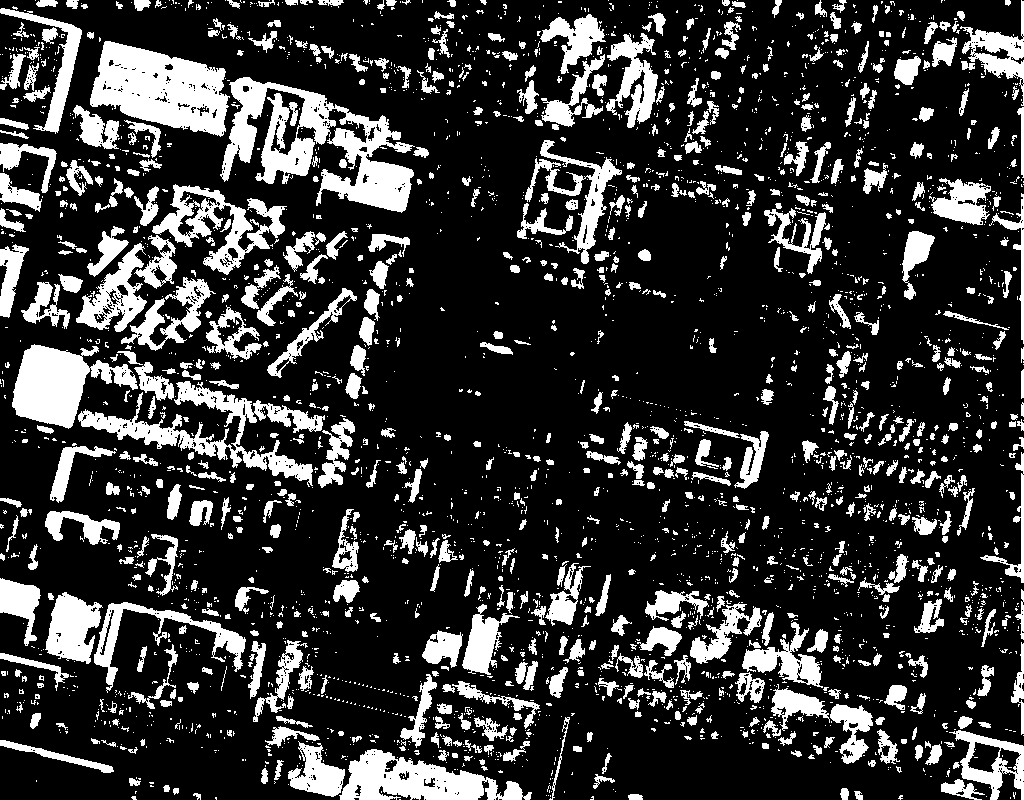}
  \label{LO_DSFANet}}
  \hfil
  \subfloat[]{
    \includegraphics[width=1.25in]{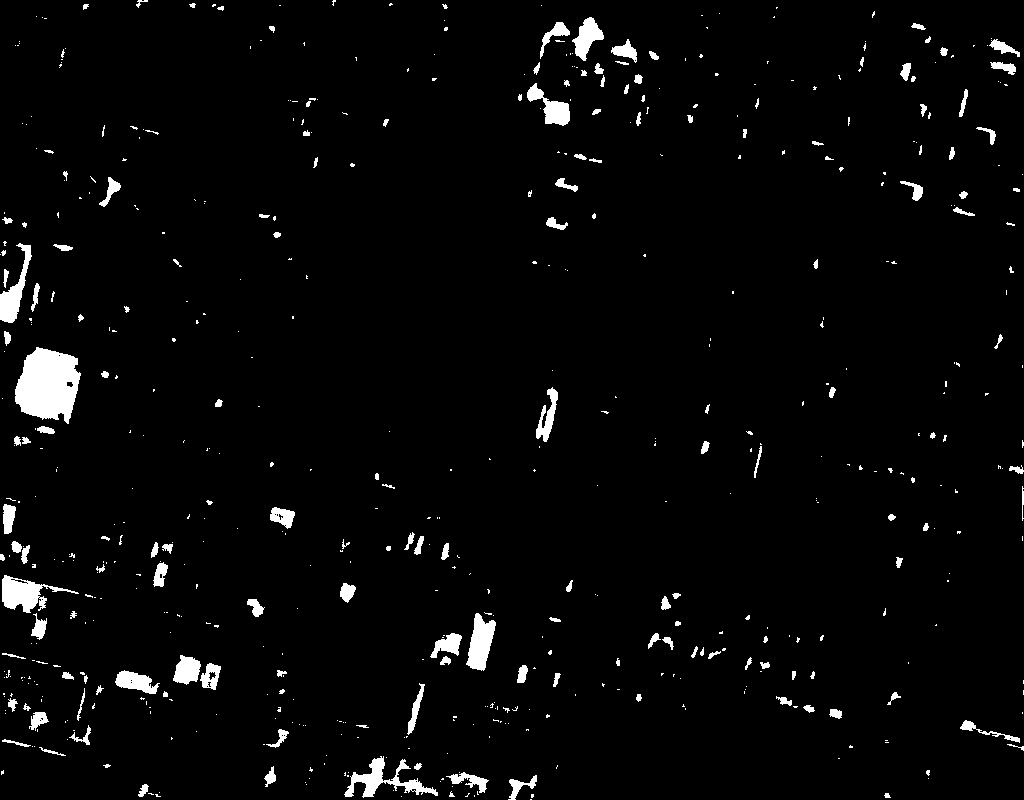}
  \label{LO_DSCNet}}
  \hfil
  \subfloat[]{
    \includegraphics[width=1.25in]{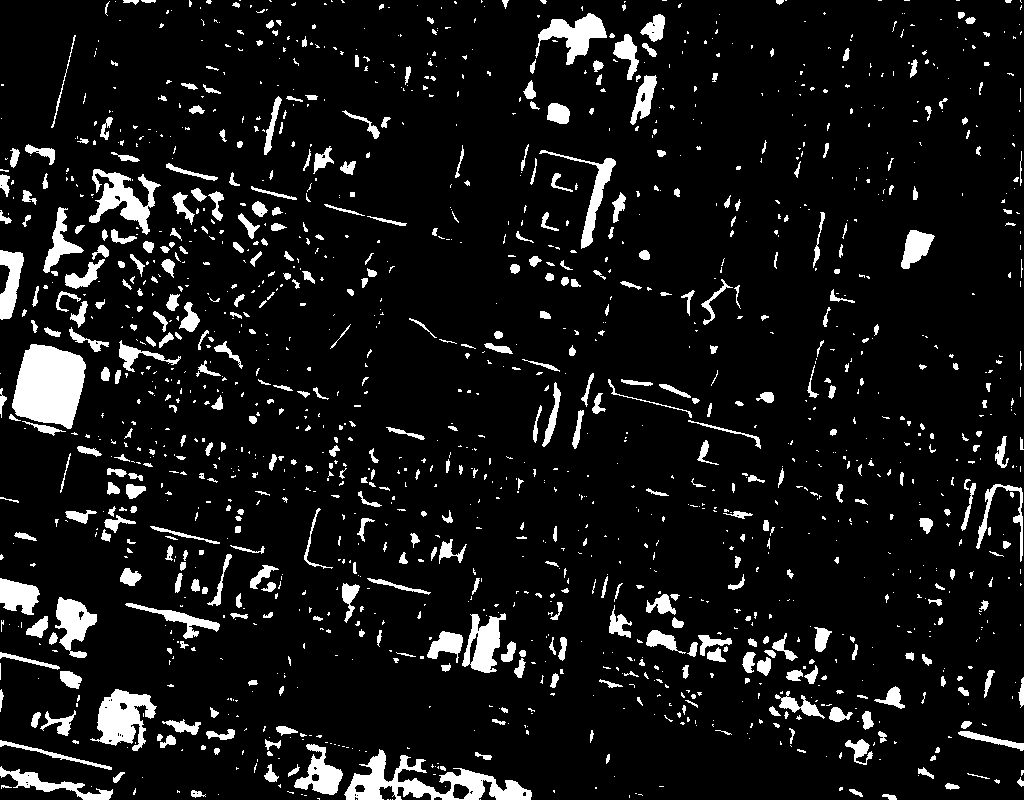}
  \label{LO_DSDANet_SK}}
  \hfil
  \subfloat[]{
    \includegraphics[width=1.25in]{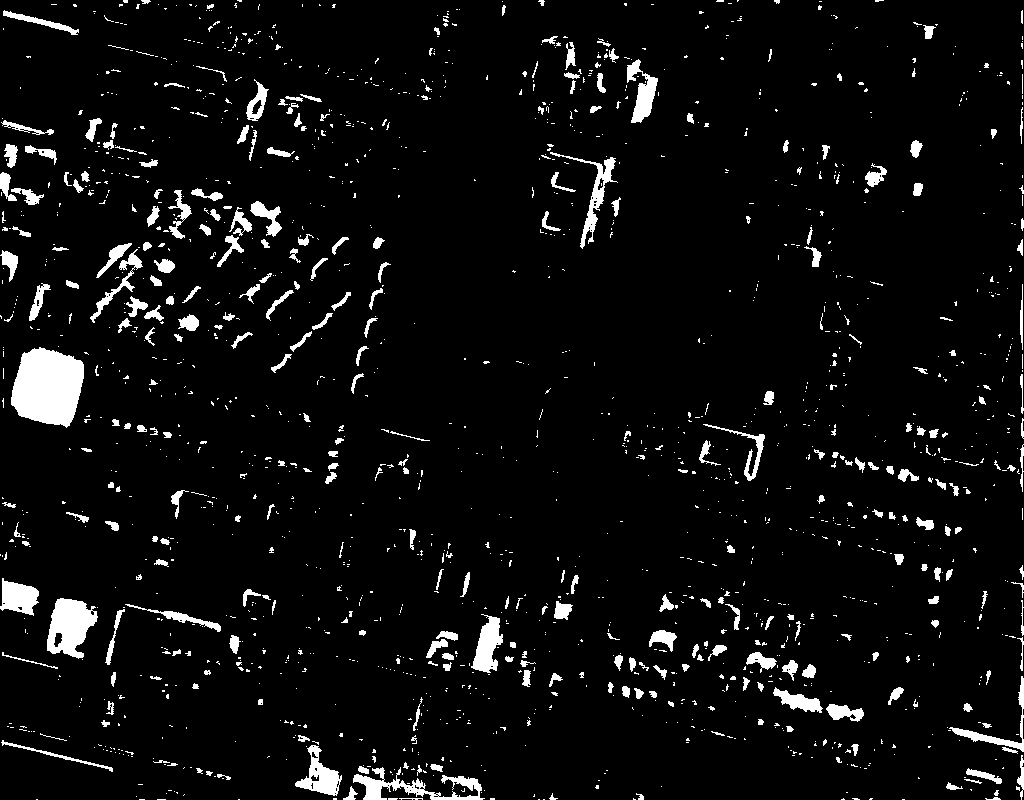}
  \label{LO_DSDANet_SL}}
  \hfil
  \subfloat[]{
    \includegraphics[width=1.25in]{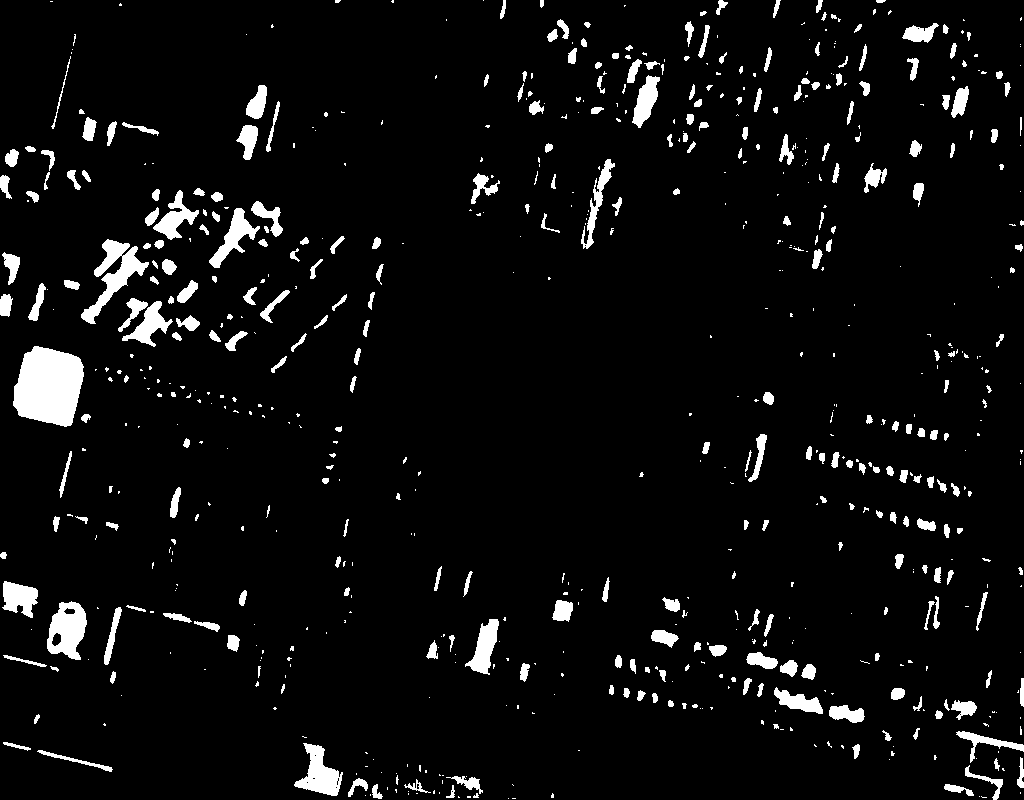}
  \label{LO_DSDANet}}

  \caption{Binary change maps obtained by different methods on the LO data set. (a) IRMAD. (b) CVA. (c) OBCD. (d) SVM. (e) RNN. (f) DSFANet. (g) DSCNet. (h) DSDANet-SK. (i) DSDANet-SL. (j) DSDANet. }
  \label{LO_result}
\end{figure*}

 On the WH$\rightarrow$HY task, because the source domain and target domain were acquired by the same sensor, leading to slightly similar distributions of two domains, thus the change map obtained by DSCNet is decent. As shown in Fig. \ref{HY_result}-(g), the main changes of bare soil, building, and roads are correctly classified. However, numerous unchanged pixels are misclassified as change class, resulting in a high FP and OE, which indicates that fine-tuning without domain adaptation performs not well in the case of providing sparse target labeled data, even if the discrepancy between two domains is not large. On the latter two data sets, due to the quite different distributions of source domain and target domain, the performance of DSCNet on the two tasks is unsatisfactory. In Fig. \ref{QU_result}-(g) and Fig. \ref{LO_result}-(g), we can observe that a large number of changed pixels are not detected. As listed in Table \ref{QU_table} and Table \ref{LO_table}, DSCNet only gets OA of 80.70$\%$ and 90.71$\%$, and KC of 0.4992 and 0.5074. Therefore, if the domain discrepancy is great, the knowledge learned by deep networks from the source domain can hardly be transferred to the target domain.

\begin{table}[t]
  \footnotesize 
  \centering
  \renewcommand{\arraystretch}{1.3}
  \caption{ACCURACY ASSESSMENT ON CHANGE DETECTION RESULTS OBTAINED BY DIFFERENT METHODS ON QU DATA SET.}
  \label{QU_table}
  \begin{tabular}{c c c c c c}
    \hline
    \bfseries Method & \bfseries FP & \bfseries FN & \bfseries OE & \bfseries OA & \bfseries KC\\
    \hline\hline
    IRMAD	& 2746	& \underline{2166}	& 4912	& 83.82	& 0.6285 \\
    CVA	& 2310	& 3521	& 5831	& 80.79	& 0.5352 \\ 					
    OBCD	& \underline{709}	& 4759	& 5468	& 81.99	& 0.5233 \\ 					
    SVM & \textbf{231}	& 4575	& 4806	& 84.17 &	0.5769 \\ 					
    RNN	& 4684	& 3431	& 8115	& 73.27	& 0.3979 \\ 				
    DSFANet	& 2655	& \textbf{2105}	& 4760	& 84.32	&	0.6398 \\ 
    DSCNet	& 1200	& 4657	& 5857	& 80.70	& 0.4992 \\ 		
    \hline			
    DSDANet-SK	& 1269	& 2657	& \underline{3926}	& \underline{87.07}	& \underline{0.6854} \\ 		
    DSDANet-SL	& 1987	& 2217	& 4204	& 86.15	& 0.6745 \\
    DSDANet	& 858	& 2866	& \textbf{3724}	& \textbf{87.73}	& \textbf{0.6958} \\ 					
    \hline
  \end{tabular}
\end{table}

\par Compared to DSCNet, the proposed method adopting domain adaptation generates obviously better CD results. On the HY data set, DSDANet is capable of achieving accuracy increments of 3.05$\%$ and 0.1004 for OA and KC, respectively; On the QU data set, improvements in OA and KC achieved by DSDANet are 7.03$\%$ and 0.1966, respectively, On the last data set, the accuracy increments on OA and KC are, respectively, 3.93$\%$, and 0.2332. Owing to the larger domain discrepancy on the latter two tasks, the accuracy improvement brought by domain adaptation is higher. In addition, the two variants of DSDANet, i.e. DSDANet-SK and DSDANet-SL can also achieve significant accuracy increments on the three tasks. Consequently, through minimizing domain discrepancy with conventional CD loss together, the proposed model can learn transferable difference feature representation, thereby successfully using the knowledge of source domain to solve the target tasks. 

\subsubsection{Different Domain Adaptation Schema}
\par To further prove the effectiveness of the proposed domain adaptation strategy, we compared DSDANet with its two variants, i.e. DSDANet-SK and DSDANet-SL. As shown in Table \ref{HY_table}-Table \ref{LO_table}, on the three tasks, DSDANet outperforms DSDANet-SK on OE, OA, and KC. For example, on the LO data set, in comparison with DSDANet-SK, DSDANet improves the accuracy by 0.48$\%$ of OA and 0.0069 of KC. The experimental results reveal the fact that adopting multi-kernels can find a more optimal RHKS and have better domain adaptation effect, thus improving the transferability of learned difference features. For another variant DSDANet-SL, on the HY data set, owing to the similar distribution, DSDANet doesn’t show obvious superiority. However, on both QU and LO data sets, DSDANet generates better change maps and outperforms DSDANet-SL on FP, OE, OA, and KC, which demonstrate the necessity of adaptation in multiple layers, especially in the case of large domain discrepancy. Therefore, by jointly exploring MK-MMD with multi-layer adaptation schema, the proposed DSDANet can learn more transferable feature representation, yielding better CD results.

\begin{table}[t]
  \footnotesize 
  \centering
  \renewcommand{\arraystretch}{1.3}
  \caption{ACCURACY ASSESSMENT ON CHANGE DETECTION RESULTS OBTAINED BY DIFFERENT METHODS ON LO DATA SET.}
  \label{LO_table}
  \begin{tabular}{c c c c c c}
    \hline
    \bfseries Method & \bfseries FP & \bfseries FN & \bfseries OE & \bfseries OA & \bfseries KC\\
    \hline\hline
    IRMAD	& 2362	& 7384	& 9746	& 91.67 & 0.6770 \\
    CVA	& 5840	& 8782	& 14622	& 87.51	& 0.4893 \\ 					
    OBCD	& 5123	& 5508	& 10631	& 90.92	& 0.5935 \\ 					
    SVM & 3592	& 3992	& 7584	& 93.52 &	0.7099 \\ 					
    RNN	& 4641	& \underline{3529}	& 8170	& 93.02 & 0.7003 \\ 				
    DSFANet	& 11940	& \textbf{1005}	& 12945	& 88.94 &	0.6252 \\ 
    DSCNet	& 2492	& 8384	& 10876	& 90.71 & 0.5074 \\ 		
    \hline			
    DSDANet-SK	& 2879	& 3951	& 6830	& 94.16	& 0.7337 \\ 		
    DSDANet-SL	& \underline{2015}	& 4437	& \underline{6452}	& \underline{94.49}	& \underline{0.7382} \\
    DSDANet	& \textbf{1625}	& 4652	& \textbf{6277}	& \textbf{94.64}	&  \textbf{0.7406} \\ 					
    \hline
  \end{tabular}
\end{table}

\subsubsection{Other Comparison Methods}
\par For the three unsupervised CD models, though they do not require any labeled data and can be directly performed on the data sets, they cannot generate accurate change maps for all target data sets. The inability to use label information of two domains limits their accuracy. For example, IRMAD obtains relatively accurate results on QU and LO data sets with KP of 0.6285 and 0.6770, respectively. However, on the HY data set, the result of IRMAD is unsatisfactory. As shown in Fig. \ref{HY_result}-(a), numerous unchanged regions are misclassified as change class and lots of changed pixels are not detected. Compared to DSCNet, SVM and DSFANet get better results on three target data sets. SVM shows a certain degree of effectiveness in cross-domain CD, but it still cannot compete with our method. RNN yields a good result on the LO data sets with OA of 93.02$\%$ and KP of 0.7003, but do not show obvious transferability on the other two target data sets. Eventually, as observed in Table \ref{HY_table} to Table \ref{LO_table}, the proposed DSDANet achieves the best performance on three data sets, followed by its two variants.

\par In summary, all the above results reveal that through embedding data distributions into the optimal RKHS and minimizing the distance between them, DSDANet is capable of learning transferable representation from source labeled data and unlabeled target data and can be easily transferred from one CD data set to another.

\subsection{Discussion}
\subsubsection{Parameter discussion}
\begin{figure*}[!t]
  \centering
  \subfloat[]{
    \includegraphics[width=2.2in]{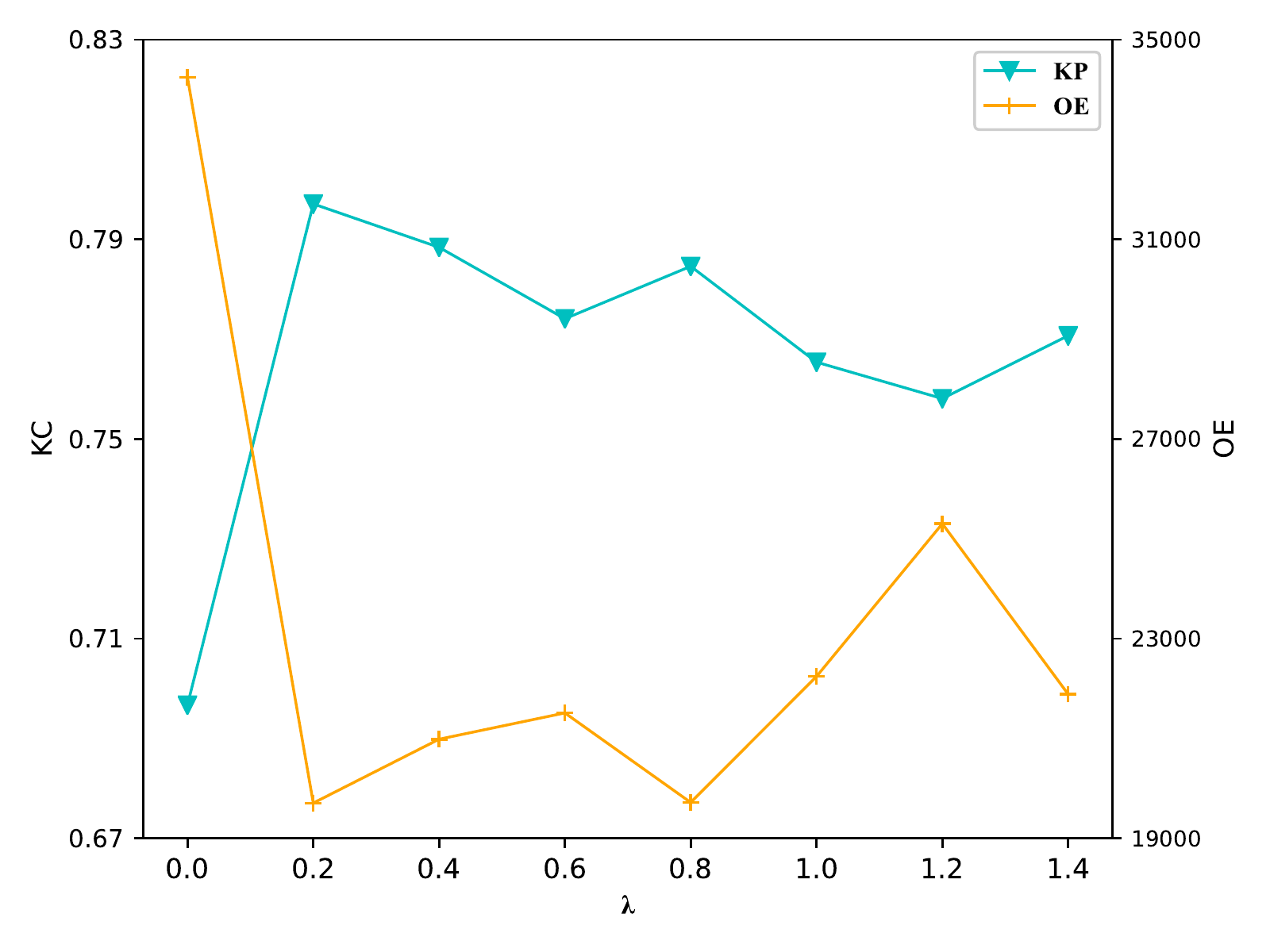}
  \label{HY_lamb}}
  \hfil
  \subfloat[]{
    \includegraphics[width=2.2in]{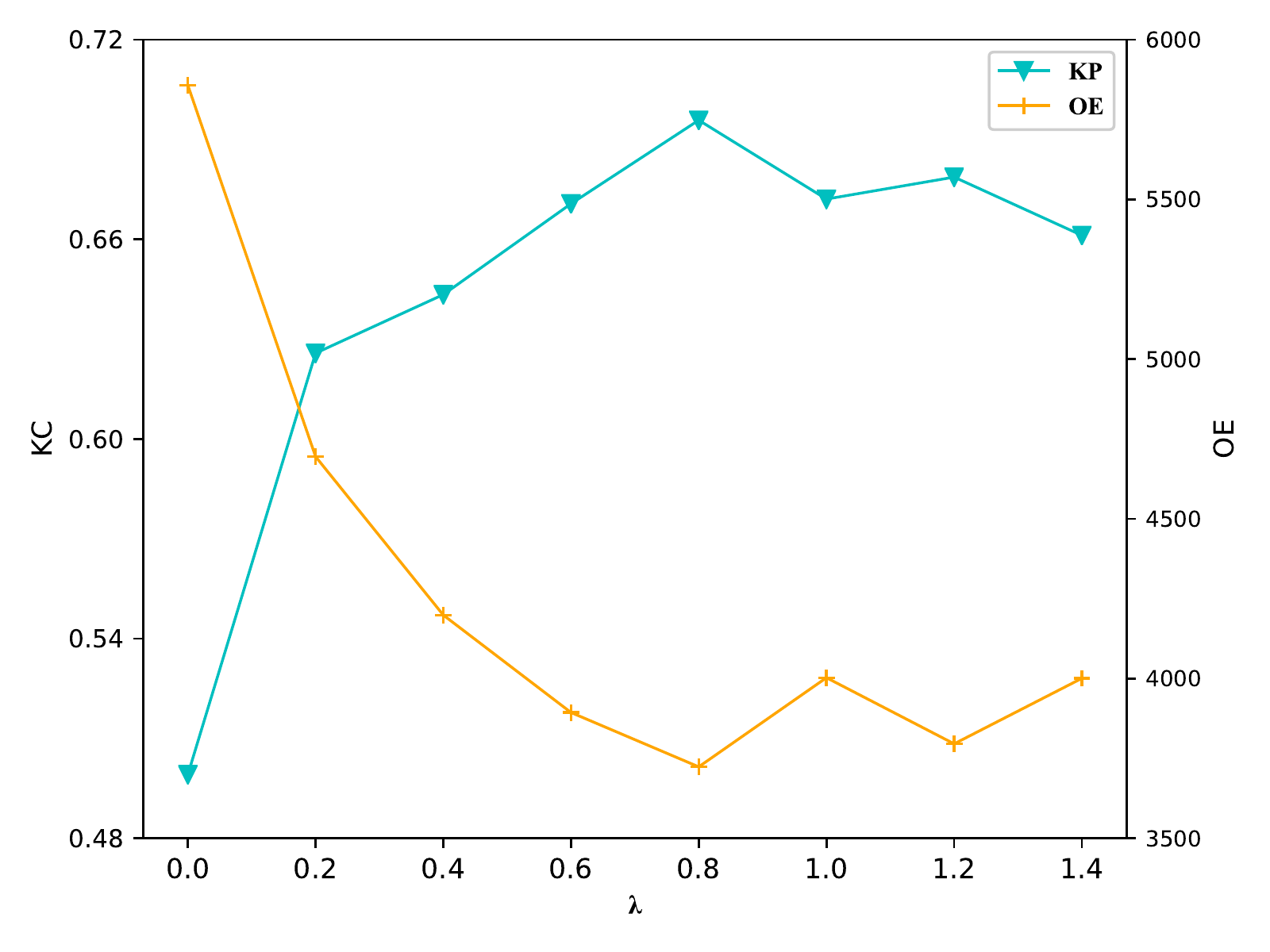}
  \label{QU_lamb}}
  \hfil
  \subfloat[]{
    \includegraphics[width=2.2in]{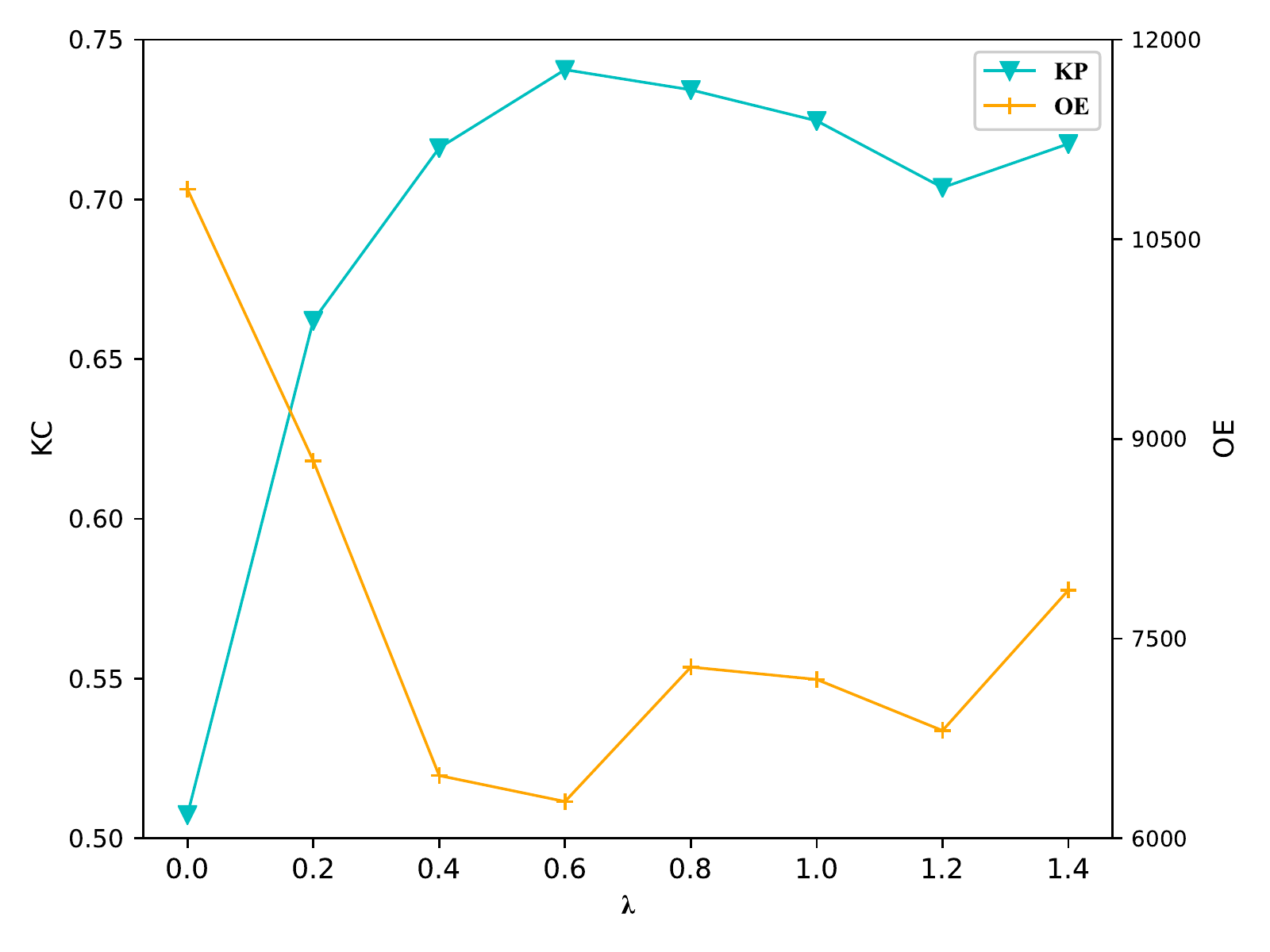}
  \label{LO_lamb}}

  \caption{Performance of the proposed DSDANet with different values of domain adaptation penalty $\lambda$. (a) HY. (b) QU. (c) LO.}
  \label{fig:dis_lamb}
\end{figure*}

\begin{figure*}[!t]
  \centering
  \subfloat[]{
    \includegraphics[width=2.2in]{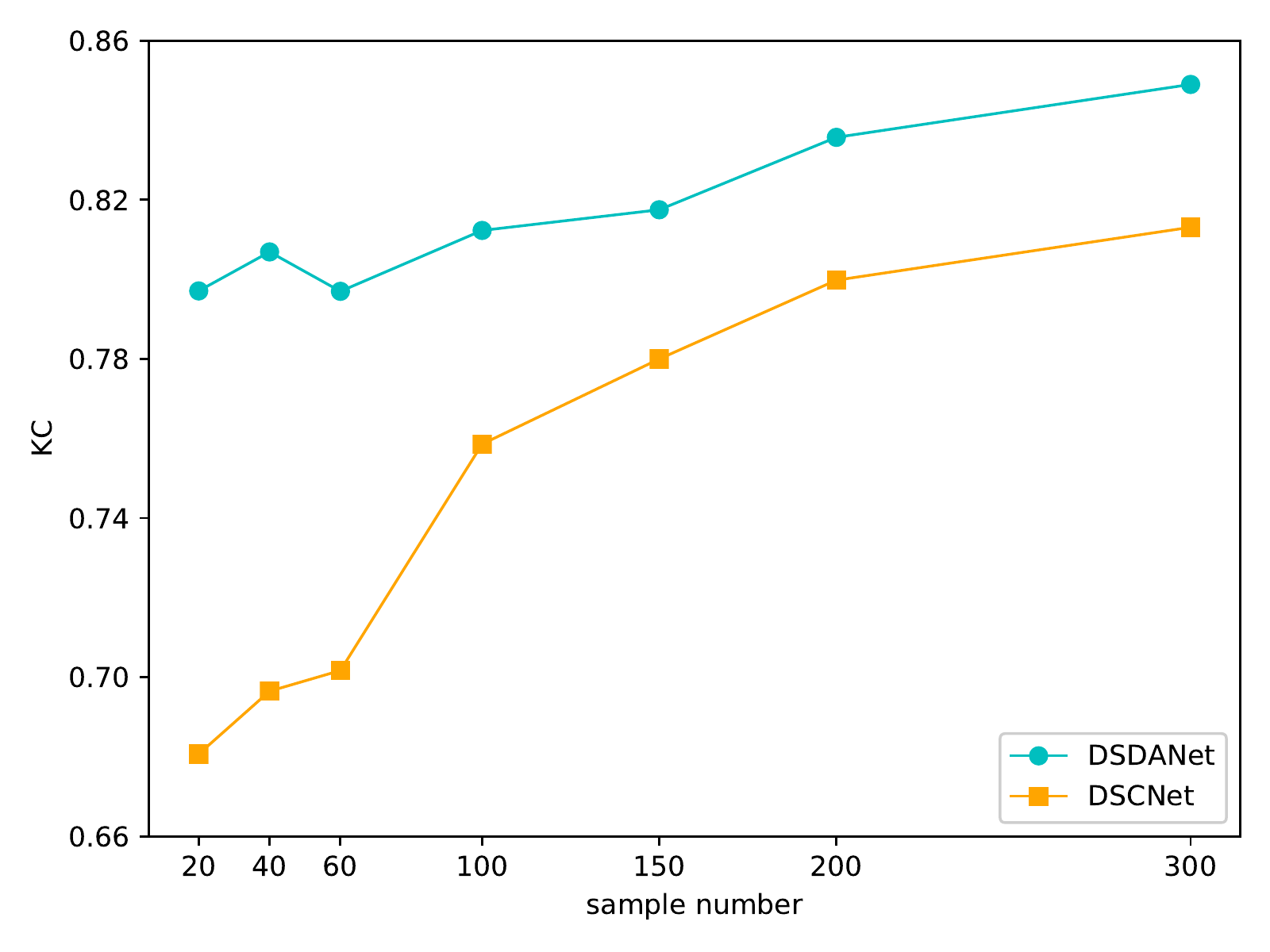}
  \label{HY_sample_num}}
  \hfil
  \subfloat[]{
    \includegraphics[width=2.2in]{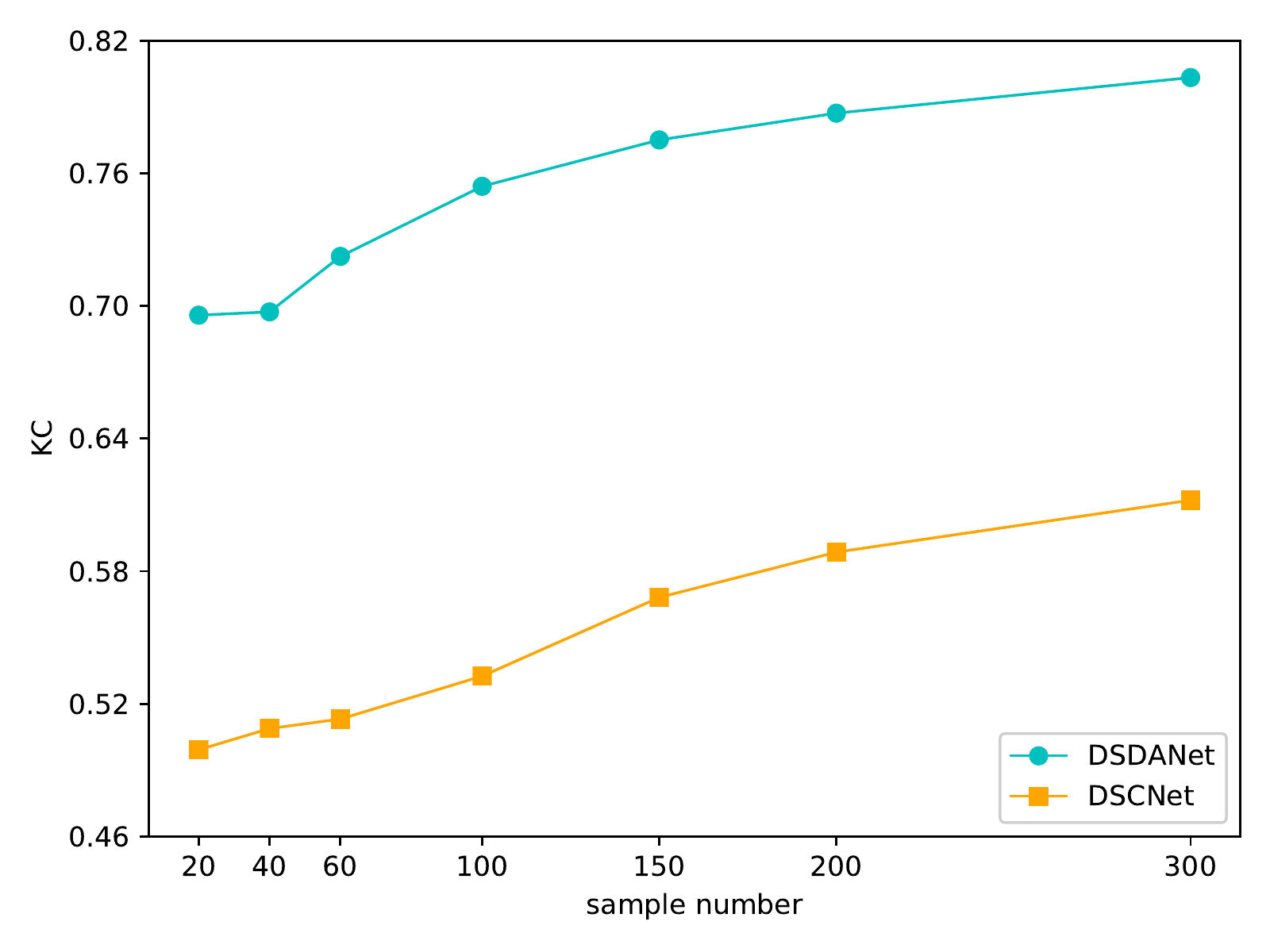}
  \label{QU_sample_num}}
  \hfil
  \subfloat[]{
    \includegraphics[width=2.2in]{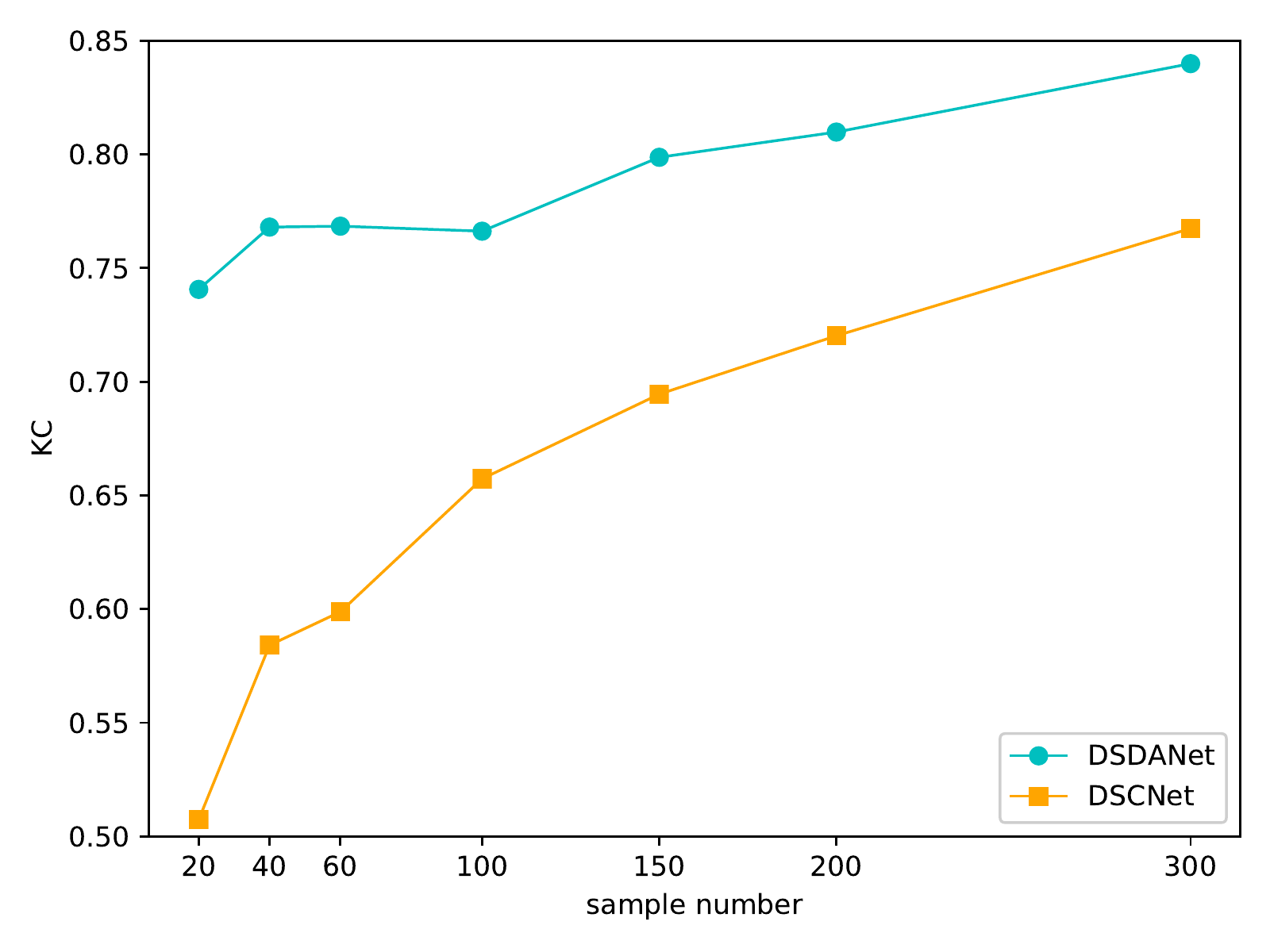}
  \label{LO_sample_num}}

  \caption{Performance of the proposed DSDANet and DSCNet with different numbers of target labeled data. (a) HY. (b) QU. (c) LO.}
  \label{fig:dis_sample_num}
\end{figure*}

\begin{figure*}[!t]
  \centering
  \subfloat[]{
    \includegraphics[width=2.2in]{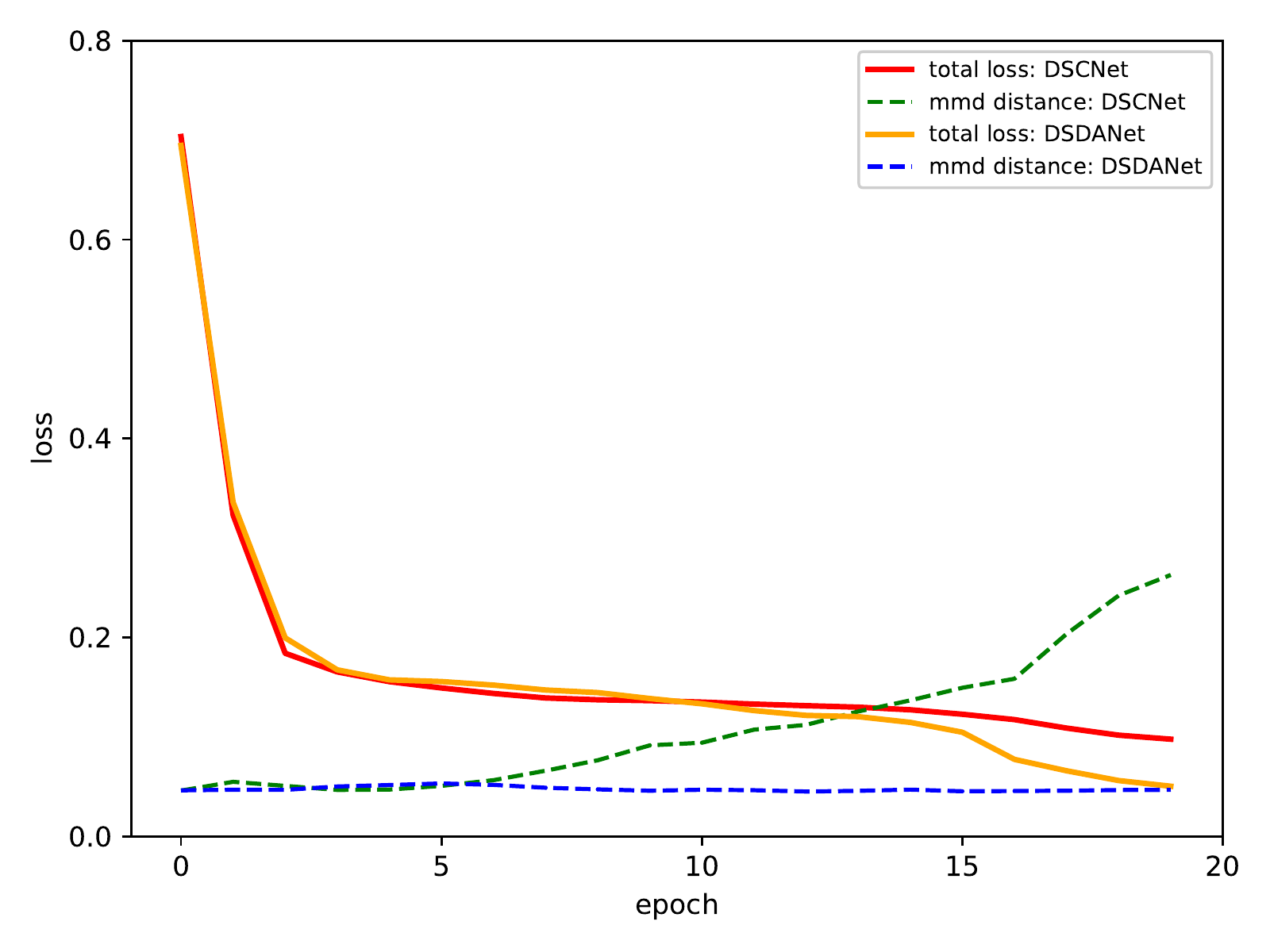}
  \label{HY_loss}}
  \hfil
  \subfloat[]{
    \includegraphics[width=2.2in]{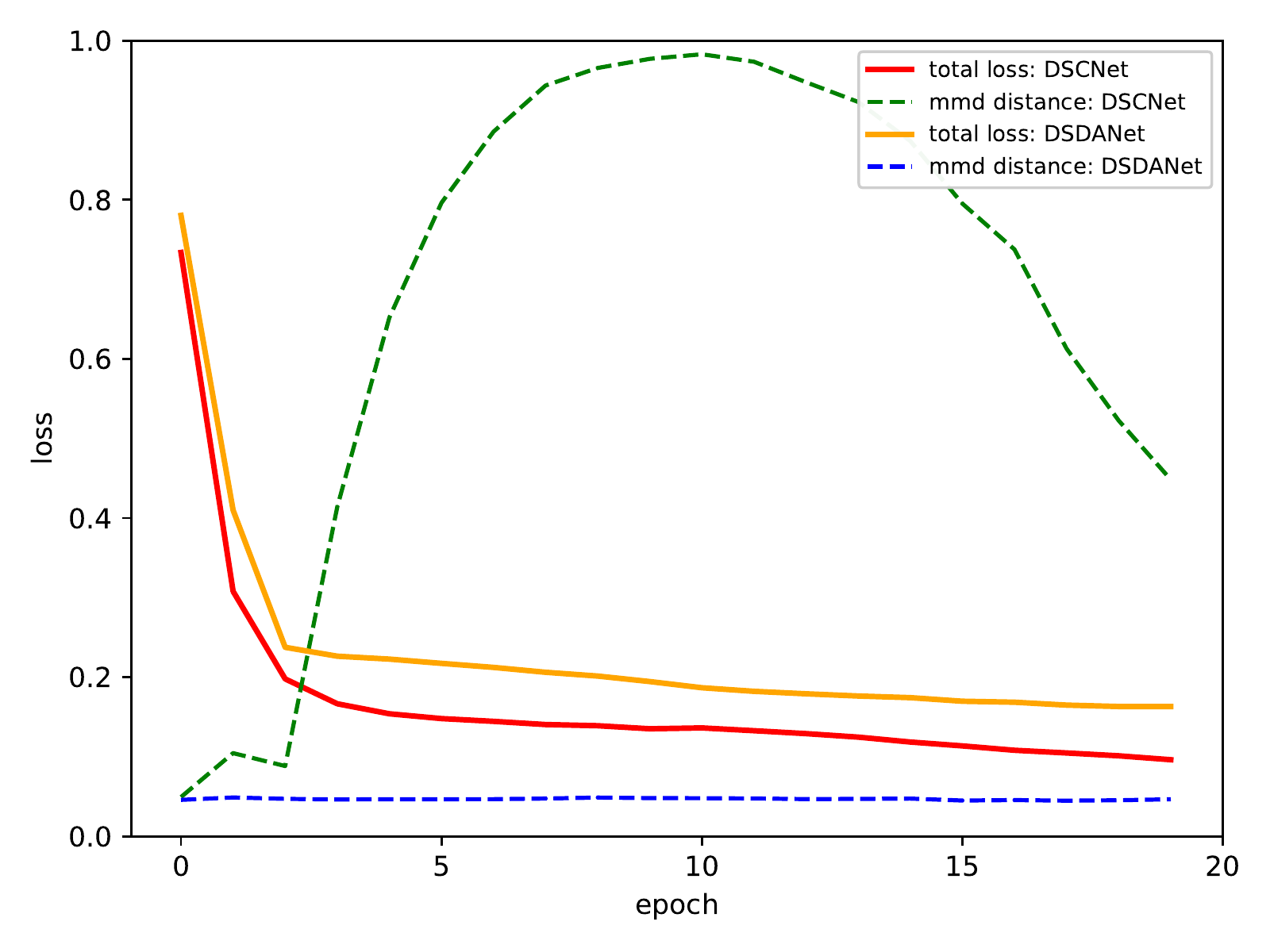}
  \label{QU_loss}}
  \hfil
  \subfloat[]{
    \includegraphics[width=2.2in]{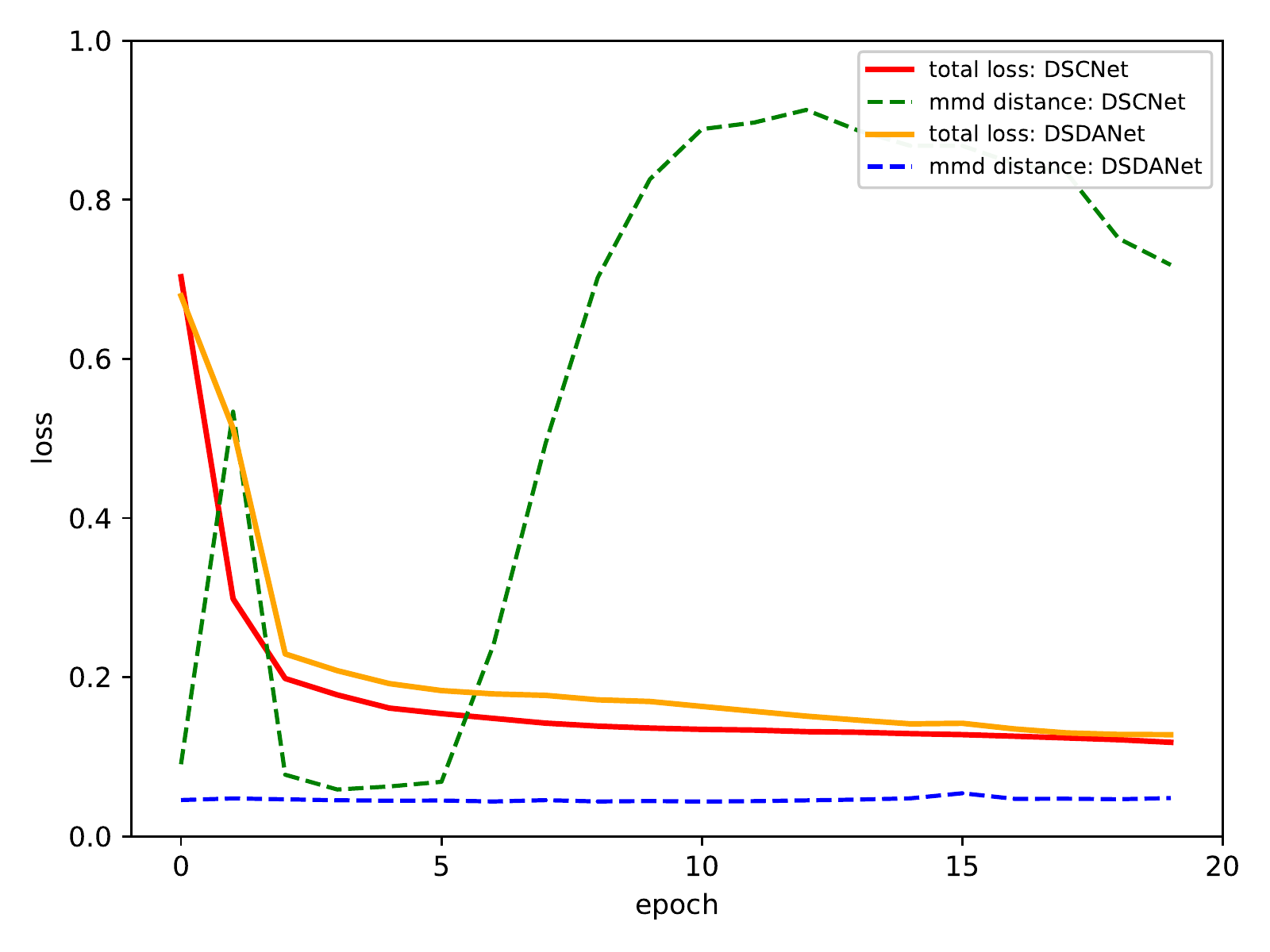}
  \label{LO_loss}}

  \caption{Loss curves of DSCNet and DSDANet on three cross-domain CD tasks. (a) WH$\rightarrow$HY. (b) WH$\rightarrow$QU. (c) WH$\rightarrow$LO.}
  \label{fig:dis_loss}
\end{figure*}

\par For our proposed DSDANet, there are two important hyper-parameters, i.e. domain adaptation penalty factor $\lambda$ and numbers of target labeled data for fine-tuning. The relationship between the CD performance of DSDANet and $\lambda$ is shown in Fig. \ref{fig:dis_lamb}. We set $\lambda$ as 0 to 1.4 with a step of 0.2, when $\lambda=0$, DSDANet degenerates to DSCNet. First, we can observe that when $\lambda$ increases from 0 to 0.2, the accuracy of model increases dramatically on all three target data sets. This confirms the effectiveness of our motivation that jointly learning deep difference features and minimizing domain discrepancy. Then the accuracy of DSDANet continues to increase and then decreases as $\lambda$ varies. Therefore, a good trade-off between conventional CD loss and domain distribution discrepancy is also important that can enhance feature transferability. Specifically, the optimal $\lambda$ for three target domains is 0.2, 0.8, and 0.6, respectively. 

\par As we mentioned, we train the DSDANet on the source labeled data first. Then we use very sparse target labeled data to fine-tune our model, the specific number is 20. In Fig. \ref{fig:dis_sample_num}, we further discuss the relationship between the accuracy of DSDANet and DSCNet and the numbers of target labeled data. It can be observed that as the increase of the number of target labeled data, more knowledge of target domain can be considered, hence the accuracy of DSDANet increases. Furthermore, on the first and third target data sets, as the number of target labeled data varies, the DSCNet becomes to approach the performance of DSDANet. But on the QU data set, the precision gap between the two methods has never been narrowed. Even if the number of target labeled data reaches to 300, the KC of DSCNet is still only 0.6121, which confirms that the performance of standard fine-tune would be seriously damaged by the over-large domain distribution discrepancy. Besides, note that the slope of accuracy curves gradually becomes smaller. This is because the backbone of DSDANet is just a conventional CNN that does not contain any advanced modules, such as multi-scale features \cite{Szegedy2015a}, residual learning \cite{He2016}, attention mechanism \cite{Zhang2018}, and so on, which limits its learning ability. Actually, the proposed domain adaptation strategy can be treated as a flexible way and used in more sophisticated networks for cross-domain CD.

\subsubsection{Learning Transferable features}
\par To dive into the proposed DSDANet and further prove its transferability, we also conduct additional experiments. First, Fig. \ref{fig:dis_loss} displays loss curves of DSDANet and DSDANet on the three tasks during training phase. On the WH$\rightarrow$HY tasks, the loss of DSCNet gradually becomes small, but the MK-MMD between two domains in the learned features gradually becomes large. On the WH$\rightarrow$QU and WH$\rightarrow$LO, for DSCNet, the distance between two domains significantly increases at begin, then drops and finally will steadily keep a high value. These facts indicate that during training phase, the features learned by DSCNet are more and more suitable for source domain, namely task-specific, but the distance between source domain and target domain becomes greater, resulting in a weak transferability of DSCNet. By contrast, as the total loss of DSDANet decreased, MK-MMD between two domains always keeps a low value, which implies that DSDANet can keep a very small domain distribution discrepancy while learning representative deep features, leading to a great transferability.

\begin{figure*}[!t]
  \centering
  \includegraphics[width=3.3in]{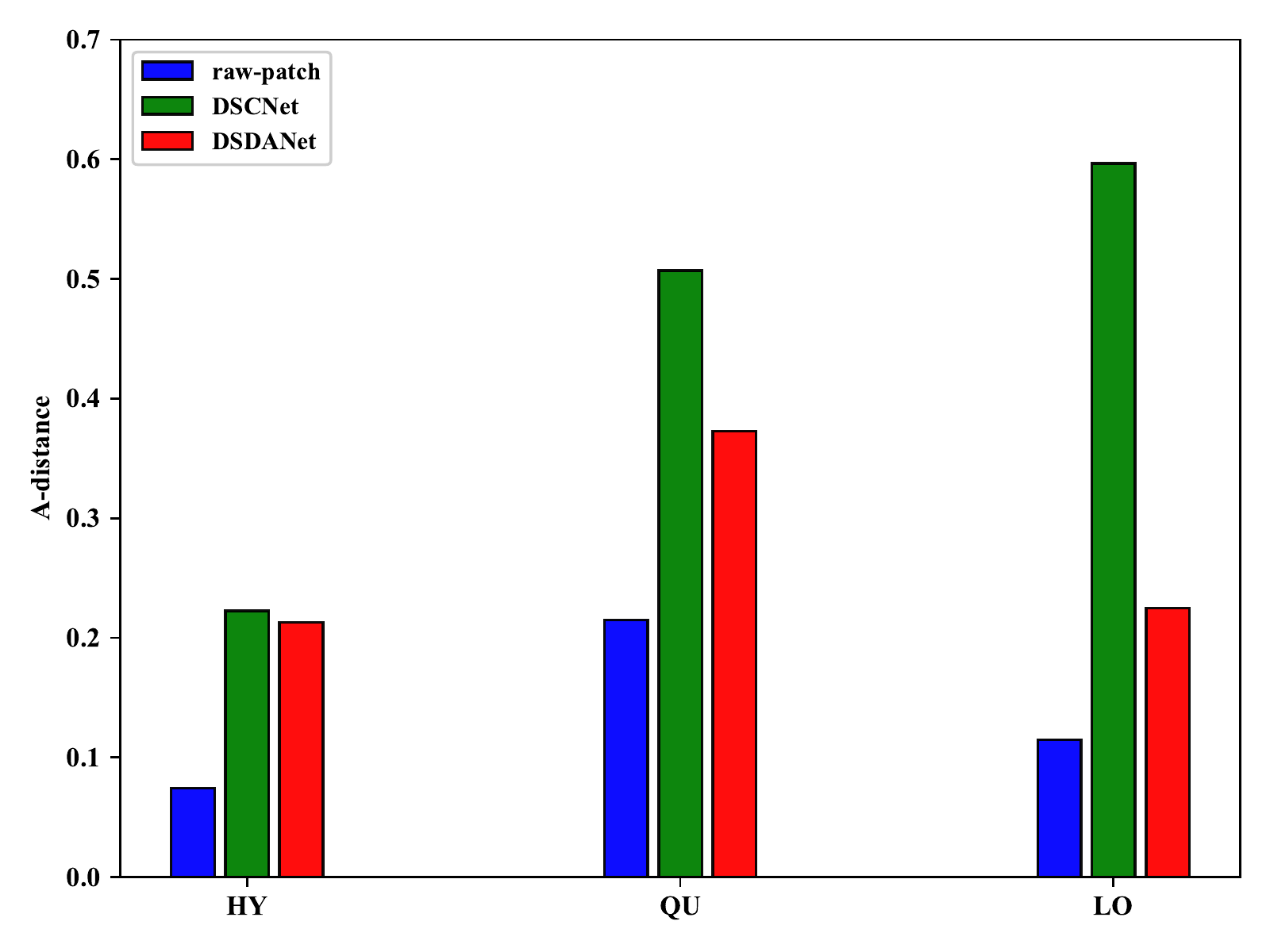}
  \caption{$\mathcal{A}$-Distance of DSCNet and DSDANet features.}
  \label{fig:A_dis}
\end{figure*}

\begin{figure*}[!t]
  \centering
  \subfloat[]{
    \includegraphics[width=1.6in]{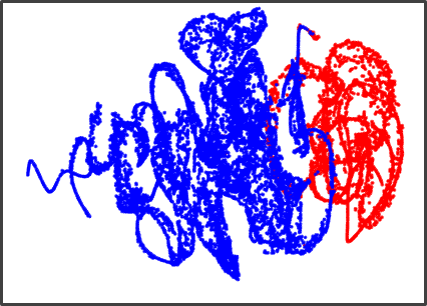}
  \label{DSCNet_WH}}
  \hfil
  \subfloat[]{
    \includegraphics[width=1.6in]{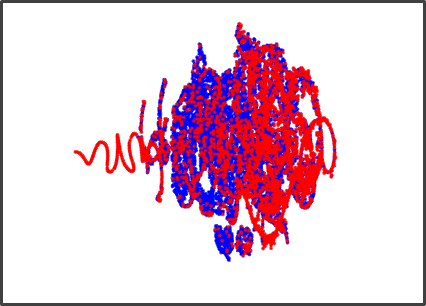}
  \label{DSCNet_QU}}
  \hfil
  \subfloat[]{
    \includegraphics[width=1.6in]{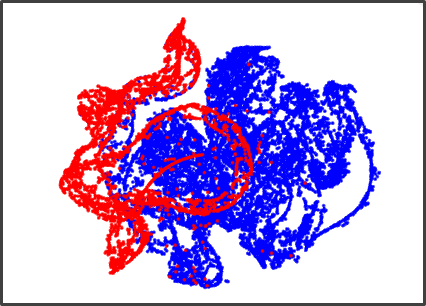}
  \label{DSDANet_WH}}
  \hfil
  \subfloat[]{
    \includegraphics[width=1.6in]{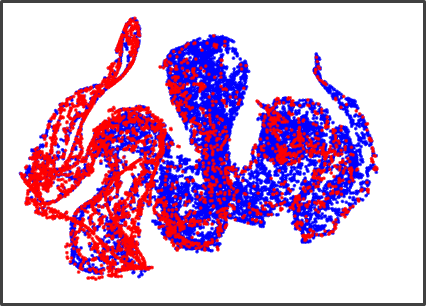}
  \label{DSDANet_QU}}

  \caption{ Visualization of DSCNet features on (a) source domain and (b) target domain and DSDANet features on (c) source domain and (d) target domain on the task WH$\rightarrow$QU by t-SNE. Red indicates change class and blue means non-change class. }
  \label{fig:tSNE}
\end{figure*}

\par In transfer learning, $\mathcal{A}$-distance is a common measurement that is often adopted to represent the similarity between the distributions of two domains \cite{Ben-David2009}. To compute $\mathcal{A}$-distance, we need to construct a linear classifier $f$ (linear SVM is adopted in this paper) in the source domain $\mathcal{D}_{s}$ and target domain $\mathcal{D}_{t}$ that aims to distinguish which domain the input samples come from. Then $\mathcal{A}$-distance is defined as
\begin{equation}
  \mathcal{A}\left(\mathcal{D}_{s}, \mathcal{D}_{t}\right)=2\left(1-\mathcal{L}_{hinge}(f)\right)
\end{equation}
where $\mathcal{L}_{hinge}(f)$ is the hinge loss of the linear classifier. Thus, the larger the $\mathcal{A}$-distance between two domains, the more different the distributions of two domains. Fig. \ref{fig:A_dis} shows $\mathcal{A}$-distance on the three tasks utilizing raw image patches, deep features of DSCNet, and deep features of DSDANet. First, we can obverse that the $\mathcal{A}$-distance on raw image patches is smaller than the ones on deep features of DSCNet and DSDANet, which means that the difference features learned by deep models can both efficiently distinguish change and non-change, and simultaneously lead to more discriminative distributions of two domains. However, the $\mathcal{A}$-distance of DSDANet is obviously smaller than the ones of DSCNet, especially on the WH$\rightarrow$QU and WH$\rightarrow$LO tasks, which demonstrates that DSDANet is capable of learning more transferable difference features. 

\par Finally, we utilize t-SNE \cite{Laurens2014} algorithm to visualize the deep difference features learned by DSCNet and DSDANet on the task WH$\rightarrow$QU. As illustrated in Fig. \ref{fig:tSNE}, in the features of DSCNet, the data distributions of two domains are very different. The deep features of DSCNet are task-specific and only applicable for source domain. In the target domain, change and non-change classes are completely mixed. This result can explain the previous experimental result that even if the number of target labeled data for fine-tuning reaches to a large number, the performance of DSCNet is still unsatisfactory. In comparison with DSCNet, in the features learned by DSDANet, the data distributions of two domains are significantly similar on the whole. It proves that our proposed domain adaptation strategy can minimize domain distribution discrepancy and enhance the transferability of learned features, thus DSDANet only requires very sparse target labeled data for fine-tuning to achieve promising CD performance in target data sets.

\section{Conclusion}\label{sec:5}
\par In this paper, a novel network architecture entitled DSDANet is presented for cross-domain CD in multispectral images. By restricting the domain discrepancy with MK-MMD and optimizing the network parameters and kernel coefficient, DSDANet can learn transferrable representation from source labeled data and target unlabeled data, thereby efficiently bridging the discrepancy between two domains. The experimental results on three target data sets demonstrate the effectiveness of the proposed DSDANet in cross-domain CD. Even though the data distributions of the two domains are significantly different, DSDANet only needs sparse labeled data of the target domain for fine-tuning, which makes it superior in actual production environments. 

\par Our future work can be mainly divided into three aspects. First, the proposed method only considers marginal distributions, thus the conditional distributions of two domains are still different. In the next work, we will consider how to jointly adapt the marginal distribution and conditional distribution of two domains, thereby not requiring any prior-knowledge for target domain. Then, as we mentioned, this work only involves in the case 1 and case 2 of cross-domain CD, the challenging case 3 is not be considered. Lastly, this work focuses on binary CD, it is interesting to explore how to extend this work to multi-class CD.

% Can use something like this to put references on a page
% by themselves when using endfloat and the captionsoff option.
\ifCLASSOPTIONcaptionsoff
  \newpage
\fi

\bibliographystyle{IEEEtran}
% argument is your BibTeX string definitions and bibliography database(s)
\bibliography{DSDANet.bib}

\end{document}